\documentclass[conference]{IEEEtran}
\IEEEoverridecommandlockouts

\usepackage{cite}
\usepackage{amsmath,amssymb,amsfonts}
\usepackage{algorithmic}
\usepackage{graphicx}
\usepackage{textcomp}
\usepackage{xcolor}
\def\BibTeX{{\rm B\kern-.05em{\sc i\kern-.025em b}\kern-.08em
    T\kern-.1667em\lower.7ex\hbox{E}\kern-.125emX}}
\usepackage{soul}

\usepackage{url}
\usepackage{footmisc}
\usepackage{caption}
\usepackage{ragged2e}
\usepackage{booktabs}
\usepackage{hhline}
\usepackage{multirow}
\usepackage{makecell} 
\usepackage{placeins}

\usepackage{stfloats}  

\usepackage[colorlinks=true, linkcolor=blue, citecolor=red, urlcolor=blue]{hyperref}  

\usepackage{flushend} 

\usepackage{balance} 

\usepackage{placeins} 

\begin{document}

\title{Wage Sentiment Indices Derived from Survey Comments via Large Language Models}

\author{\IEEEauthorblockN{Taihei Sone}
\IEEEauthorblockA{\textit{College of Engineering and Applied Science} \\
\textit{University of Colorado Boulder}\\
Boulder, CO, USA \\
Taihei.Sone@colorado.edu}
}

\maketitle

\begin{abstract}
	The emergence of generative Artificial Intelligence (AI) has created new opportunities for economic text analysis. This study proposes a Wage Sentiment Index (WSI) constructed with Large Language Models (LLMs) to forecast wage dynamics in Japan. The analysis is based on the Economy Watchers Survey (EWS), a monthly survey conducted by the Cabinet Office of Japan that captures real-time economic assessments from workers in industries highly sensitive to business conditions. The WSI extends the framework of the Price Sentiment Index (PSI) used in prior studies, adapting it specifically to wage-related sentiment. To ensure scalability and adaptability, a data architecture is also developed that enables integration of additional sources such as newspapers and social media. Experimental results demonstrate that WSI models based on LLMs significantly outperform both baseline approaches and pretrained models. These findings highlight the potential of LLM-driven sentiment indices to enhance the timeliness and effectiveness of economic policy design by governments and central banks.
\end{abstract}


\section{Introduction}\label{sec:introduction}
\IEEEPARstart{M}{aintaining} price stability and sustainable economic growth requires a deep understanding of labor market dynamics, particularly the trends in wages. Wages play a critical role in shaping consumer purchasing power, inflationary pressures, and broader economic activity. In recent years, substantial research has focused on using Large Language Models (LLMs) for analyzing economic data, with much of the emphasis placed on price forecasting. However, despite the importance of wages in economic planning and policy formation, studies on wage prediction through text mining have remained scarce. Traditional methods for wage analysis have relied on structured numerical data, such as employment statistics and industry-specific reports, which have often suffered from publication delays and limited predictive power. By contrast, unstructured textual data from surveys and public sentiment can offer timely and nuanced insights into evolving wage trends.

This study aims to fill this research gap by developing a ``Wage Sentiment Index (WSI)'' using text analysis on the Economy Watchers Survey (EWS) published by the Cabinet Office of Japan~\cite{ews}. The WSI was constructed by applying the methodology used to develop the Price Sentiment Index (PSI), as presented in several previous studies~\cite{otaka2018, nakajima2021, suzuki2024, izawa2024}, to the context of wages. The EWS is a high-frequency, survey-based dataset that captures economic sentiment from various industry professionals and business owners. This study investigates the predictive capability of sentiment-derived indicators for forecasting nominal wages as reported in the Monthly Labour Survey (MLS)~\cite{mls}. Textual data from the EWS, when analyzed using LLMs, were hypothesized to provide a reliable early indicator of wage movements, allowing policymakers and businesses to react more swiftly to labor market fluctuations. In addition, with a view to incorporating other text data, such as newspapers and social media, into the analysis in the future, a data architecture designed to achieve scalability was built.

The analysis indicated that LLMs showed a tendency to outperform baseline and pretrained models in predicting wage trends with the WSIs. In the future, the aim is to develop a data pipeline capable of automatically and rapidly collecting and processing diverse and large volumes of data to forecast various economic indicators, along with Artificial Intelligence (AI) agents responsible for managing and operating it. This is intended to contribute to the timely and appropriate policy formulation by governments and central banks.

The remainder of this paper is structured as follows. Section~\ref{sec:related} reviews prior literature on text mining in economic analysis, particularly in price and wage forecasting. Section~\ref{sec:method} describes the methodology, including data collection, classification models, and the construction of the WSI. Section~\ref{sec:data} discusses the data sources used in this study. Section~\ref{sec:architecture} describes the data architecture constructed to ensure the scalability of the analysis in this study. Section~\ref{sec:results} verifies the usefulness of each WSI constructed using different models as a leading indicator of wage trends. Finally, Section~\ref{sec:conclusion} summarizes the findings of this study and suggests avenues for future research.

\section{Related Literature}\label{sec:related}
Wage prediction has been explored using Machine Learning (ML) and foundation models. In Saudi Arabia, multiple ML models, including Bayesian Gaussian Process regression, have significantly outperformed traditional linear regression~\cite{matbouli2022}. Salary forecasts for data science professionals around the world have been examined using decision trees, random forests, and gradient boosting, highlighting the effectiveness of the decision trees~\cite{arjun2024}. In the United States, the use of foundation models to estimate wage disparities has shown a 15\% improvement in wage prediction accuracy compared to standard econometric models~\cite{vafa2024}. It is evident that the application of ML and LLM has been advancing in the prediction of wage trends.

Sentiment analysis through text mining has played a crucial role in these labor market predictions. A Labor Market Conditions Index (LMCI) for China has been constructed using media sentiment, effectively tracking wage trends~\cite{bailliu2019}. Social media sentiment analysis has been applied to gauge employment concerns in China, revealing regional sentiment variations~\cite{tong2024}. Employee reviews from major tech firms have been analyzed using hybrid ML models to classify workplace satisfaction trends~\cite{gaye2021}. These approaches highlight the increasing importance of sentiment-driven analysis in labor market research.

Macroeconomic forecasting has also leveraged text mining extensively. Newspaper text analysis has improved Gross Domestic Product (GDP) and Consumer Price Index (CPI) forecasts~\cite{kalamara2022}. Sentiment indices derived from millions of news articles have demonstrated predictive power in European GDP growth~\cite{barbaglia2024}. In Malaysia, ML-based sentiment analysis has been used to forecast investment trends~\cite{ho2022}.

Based on these previous studies, sentiment analysis using ML-based or LLM-based text mining appears to be effective for predicting trends in Japanese wages. As mentioned above, although such analysis is seen in other countries, it is rarely done in Japan. Therefore, this study will address this analysis.

In this case, previous research that has conducted a similar analysis on prices rather than wages can be used as a reference. The ``Price Sentiment Index (PSI)'' has been developed using a Naive Bayes classifier on the EWS data, demonstrating its role as a leading inflation indicator~\cite{otaka2018}. Further analysis has explored PSI’s relation to macroeconomic variables~\cite{nakajima2021}. The introduction of multiple LLMs has improved inflation prediction accuracy~\cite{suzuki2024}, while refinements in LLM-based PSI models have identified shifts in inflation drivers from raw material costs to labor costs between 2022 and 2024~\cite{izawa2024}. In these previous studies, price trends were predicted using PSI constructed using ML and LLMs, and this study aims to predict wage trends by constructing a WSI using a similar approach. In fact, it has been pointed out that there is an increasing need to extend these approaches to other economic indicators beyond the inflation rate, and this study will address the issues left unresolved by previous research~\cite{suzuki2024}. Also, as the Bank of Japan, which is Japan's central bank, places great importance on whether or not a virtuous cycle of prices and wages can be confirmed in its monetary policy management, there is a particularly high need to predict trends in wages among economic indicators other than prices.

Other uses of LLMs are also progressing, such as in the analysis of the stance of the central bank's monetary policy. GPT-4 has been used to classify monetary policy stances with high accuracy~\cite{hansen2024}. Central Bank-specific LLMs (CB-LMs) have demonstrated superior performance in interpreting policy statements~\cite{gambacorta}.  A fine-tuned large language model for central bank communications (CentralBankRoBERTa) have shown high accuracy in sentiment classification for monetary policy analysis~\cite{pfeifer2023}. While the analysis in these studies is mainly for private companies to predict the monetary policy of the central bank, the analysis in this study is for the central bank to predict the trends of private companies in order to formulate monetary policy. For this reason, these previous studies and this study are looking in opposite directions.

\section{Method}\label{sec:method}
\subsection{How to Calculate the WSIs}\label{subsec:How_to_calculate_WSI}
The methodology consists of three steps. First, the EWS comments were classified into wage increase, wage decrease, and neutral wage categories using the baseline model, discriminative pretrained language models, and LLMs, with each comment assigned a probability of belonging to each category (see Subsection~\ref{subsec:How_to_classify_comments} for details of the baseline, pretrained, and LLM-based models)\footnote{When both the wage increase and wage decrease probabilities were zero, the comment was considered unrelated to wages and excluded from the WSIs calculation.}. This classification provides a structured basis for text analysis, ensuring accurate capture of key wage sentiment indicators. Second, based on the classified comments, both the standard WSI and the weighted WSI were computed using the following formulas:
\begin{equation}
	\label{eq:wsi_st}
	WSI_t^{Standard} = \frac{\alpha_t-\beta_t}{\alpha_t+\beta_t+\gamma_t} \times 100
\end{equation}
\begin{equation}
	\label{eq:wsi_wt}
	WSI_t^{Weighted} = \sum_{i=1}^{n_t}\frac{u_{i,t}-v_{i,t}}{u_{i,t}+v_{i,t}+w_{i,t}} \times 100
\end{equation}

where 

\begin{itemize}
	\item $\alpha_t$, $\beta_t$, and $\gamma_t$ are the number of wage increase comments, wage decrease comments, neutral wage comments in each month $t$, respectively.
	\item  $u_{i,t}$, $v_{i,t}$, and $w_{i,t}$ indicate the probability that the $i$-th comment in each month $t$ is related to a wage increase, a wage decrease, and a neutral wage, respectively.
	\item  $n_t$ is the total number of comments in each month $t$, which is equal to the sum of $\alpha_t$, $\beta_t$, and $\gamma_t$\footnote{Note that comments regarded as unrelated to wages were excluded.}.
\end{itemize}

These WSIs serve as a quantitative measure of wage sentiment, enabling comparisons across time periods and economic conditions. Finally, the time series of the WSIs were depicted, followed by a comparison with the nominal wage index (year-on-year) from the MLS to validate their predictive accuracy by means of Granger causality tests.

\subsection{How to Classify Comments When Calculating the WSIs}\label{subsec:How_to_classify_comments}

\subsubsection{Baseline Model}\label{subsubsec:baseline_model}

First, for each target month, all EWS records from at least two months prior were extracted, considering an approximate two-month lag between EWS reports and official wage statistics (MLS). The comments were transformed into a bag-of-words representation with standard preprocessing: lowercasing, stop word removal, and filtering to retain terms that appeared at least five times per month on average, in accordance with previous research~\cite{nakajima2021}. Second, the Pearson correlation between each term’s monthly frequency and the wage growth index was computed, following previous research~\cite{nakajima2021}. The top ten positively correlated terms and the top ten negatively correlated terms were then selected as positive words and negative words, respectively\footnote{When calculating the WSIs for a given month and for another month, the positive and negative words used for the calculation may differ, because the data period used to extract those words differs between the two cases. For example, when calculating the WSIs for December 2020, the positive and negative words are selected in advance using data up to October 2020, whereas when calculating the WSIs for December 2024, the positive and negative words are selected in advance using data up to October 2024. This makes it possible to calculate the WSIs while taking into account the dynamic changes in words related to wage trends.}. Finally, the standard WSI and the weighted WSI for the month were calculated based on the occurrence frequency and occurrence ratio of those positive and negative words in the EWS comments for the month.

\subsubsection{Discriminative Pretrained Language Models}\label{subsubsec:discriminative_pretrained_language_models}

FinBERT and DeBERTaV3 were used to classify comments~\cite{finbert2020, he2021deberta}. These models were not fine-tuned; instead, the general-purpose versions were applied as they were.

\subsubsection{LLMs}\label{subsubsec:LLMs}

GPT-3.5 Turbo, Claude 3.5 Sonnet, Gemini 1.5 Pro, Llama 3.3 70B\footnote{If an error occurred, Llama 3.1 8B was used instead~\cite{meta_llama31_8b}.}, and DeepSeek-V3-0324 were used to classify comments~\cite{openai_gpt35_turbo, anthropic_claude35, google_gemini15pro, meta_llama33_70b, deepseek_chat}. These models were not fine-tuned; instead, the general-purpose versions were applied as they were.

\section{Data}\label{sec:data}
\subsection{The EWS}\label{subsec:ews}
The EWS, published by the Cabinet Office of Japan, assesses economic conditions based on responses from business managers and employees~\cite{ews}. It is conducted monthly and is timely, with results published about two weeks after the survey month. This dataset is valuable because it captures real-time sentiment from individuals who are directly involved in economic activities, providing a leading indicator of labor market trends. Data are available from January 2000. Since the EWS responses are mainly recorded in Japanese, the DeepL API was used to translate them into English for analysis~\cite{deepl_api}.

\begin{figure}[htbp]
	\centering
	\includegraphics[width=\linewidth]{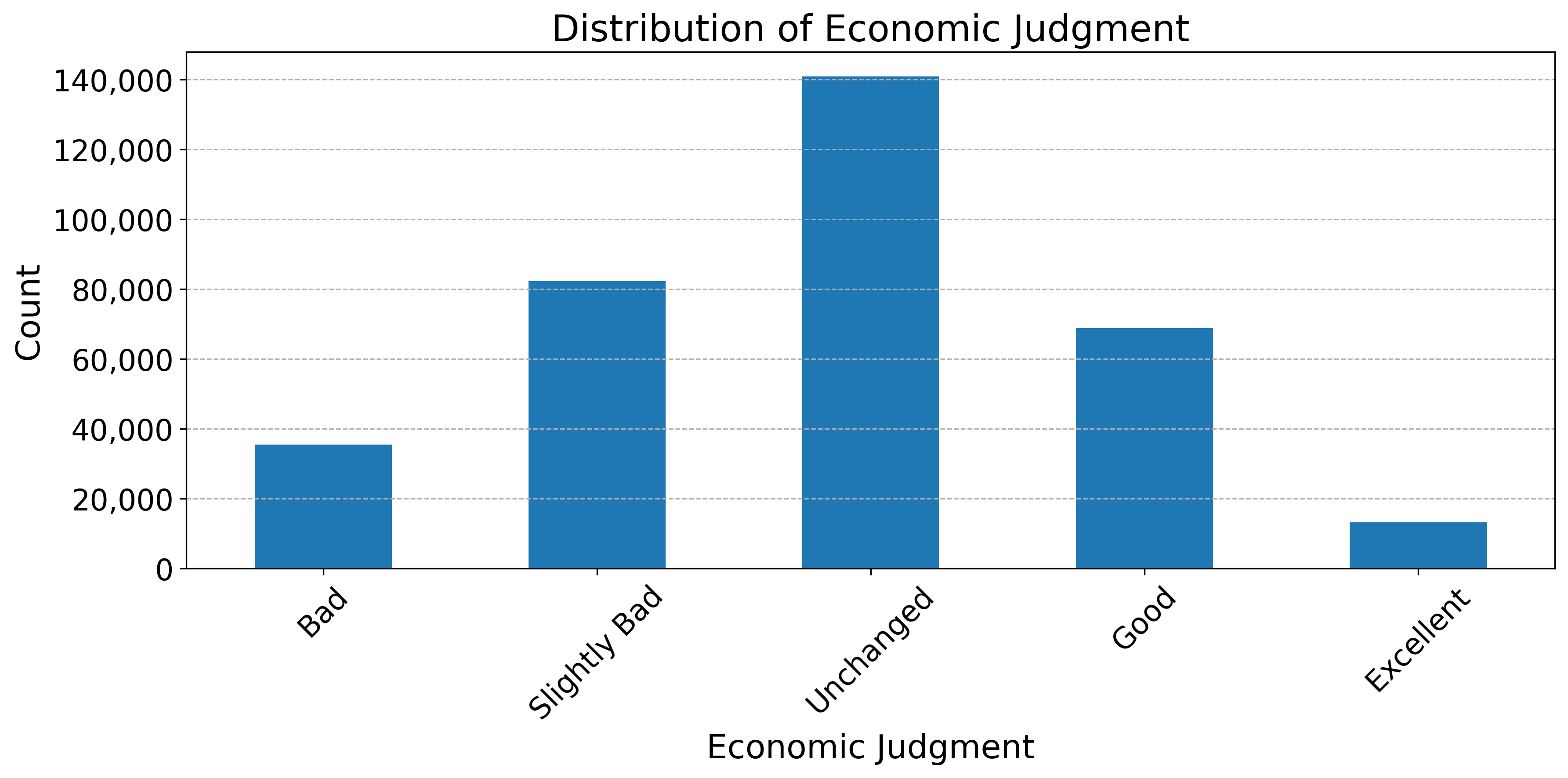}
	\caption{Distribution of Economic Judgment}
	\caption*{This figure illustrates the frequency distribution of the current economic assessment from the EWS survey conducted across Japan between January 2000 and February 2025.}
	\label{fig:05_EWS_EDA_dist_of_economic_judgment}
\end{figure}

Fig.~\ref{fig:05_EWS_EDA_dist_of_economic_judgment} presents the distribution of judgments in the responses to the EWS from January 2000 to February 2025. The response ``Unchanged'' appears most frequently, while the persistent Japanese recession over the past several decades is reflected in a greater prevalence of ``Slightly Bad'' and ``Bad'' compared to ``Excellent'' and ``Good.''

\begin{figure}[htbp]
	\centering
	\includegraphics[width=\linewidth]{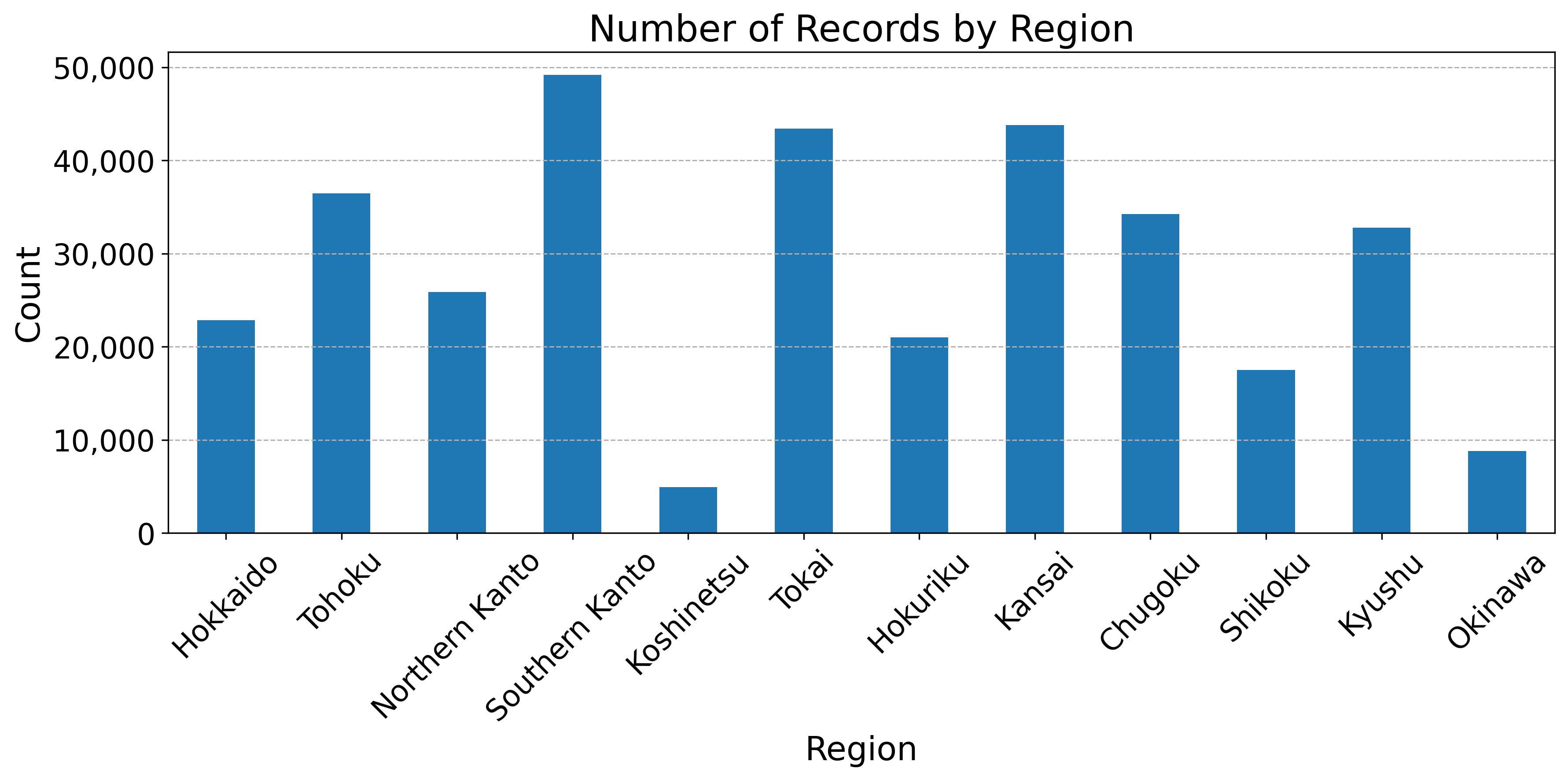}
	\caption{Number of Records by Region}
	\caption*{This figure illustrates the frequency distribution of regional response counts in the EWS survey conducted across Japan between January 2000 and February 2025.}
	\label{fig:05_EWS_EDA_n_records_by_region}
\end{figure}

\begin{figure}[htbp]
	\centering
	\includegraphics[width=\linewidth]{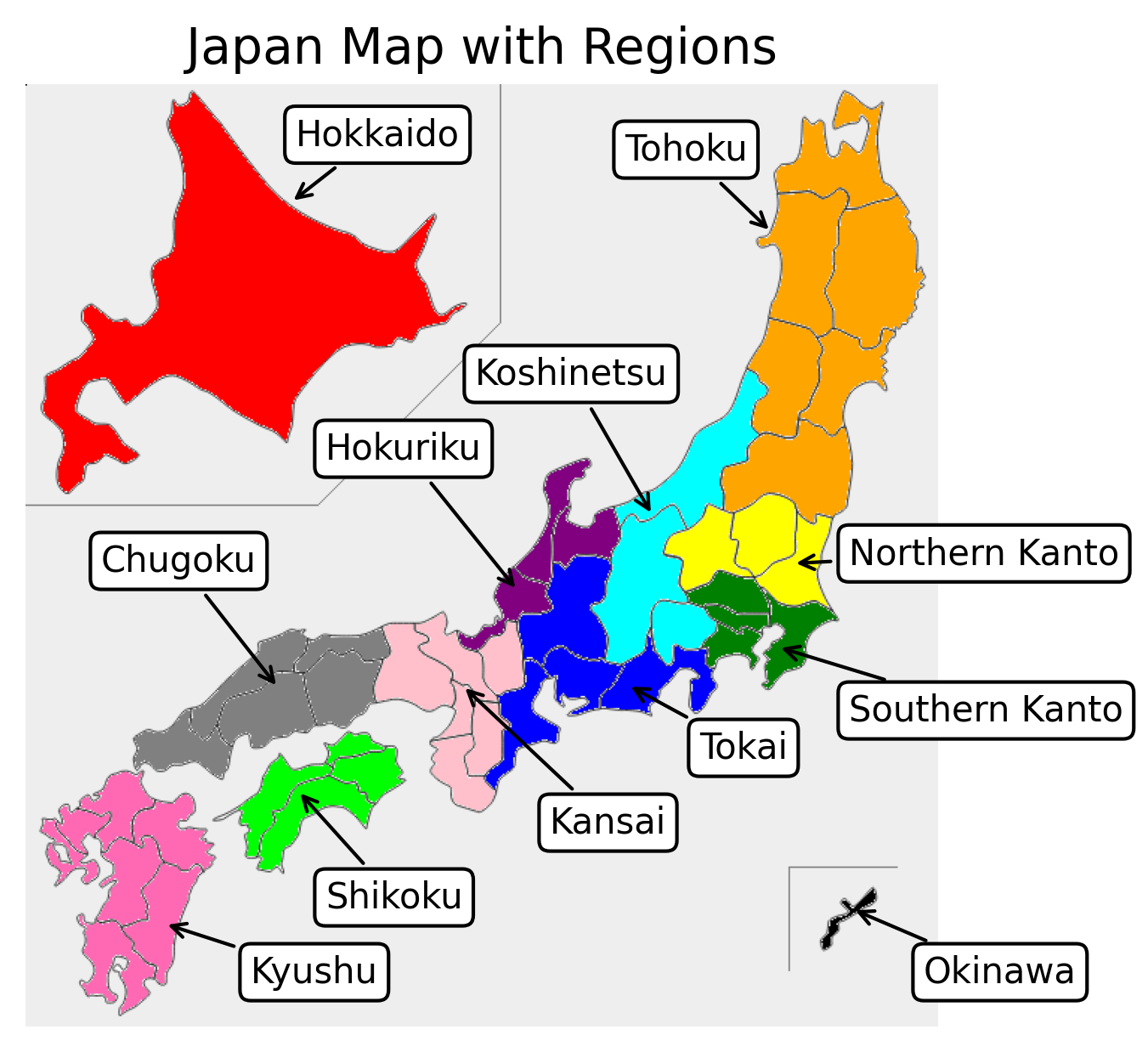}
	\caption{Japan Map with Regions}
	\caption*{This figure shows the regional divisions of Japan.}
	\label{fig:05_EWS_EDA_japan_map}
\end{figure}

Fig.~\ref{fig:05_EWS_EDA_n_records_by_region} shows the number of responses to the EWS by region from January 2000 to February 2025. This shows that there are particularly many responses from ``Southern Kanto,'' ``Tokai,'' and ``Kansai,'' which are areas with a high population density and are the economic centers of Japan. For the locations of each region, please refer to Fig.~\ref{fig:05_EWS_EDA_japan_map}.

Fig.~\ref{fig:05_EWS_EDA_monthly_judgment_trend} shows the monthly series of the total number of responses to the EWS from January 2000 to February 2025, along with a breakdown of the number of responses for each judgment. An inspection of the figure indicates an increase in the total number of responses from 2000 to around 2008–2009. This likely reflects the impact of the amendment to Japan's Statistics Act\footnote{Japan's Statistics Act (Act No. 53 of 2007) was promulgated in 2007 and entered into force in 2009. The Act stipulates fundamental matters concerning official statistics compiled by administrative organs and related bodies. It classifies official statistics into two categories—fundamental statistics and general statistics—under which reporting by survey subjects is mandatory for the former but not for the latter. As the EWS is classified as general statistics under the Act, respondents are not legally obliged to report; nevertheless, in response to this legal revision, it appears that the survey population became motivated to respond as a precaution.}. The breakdown of the responses shows that the composition ratio of ``Slightly Bad'' and ``Bad'' responses has remained relatively high, reflecting the decades-long recession in Japan. In particular, the composition ratio of these responses rose sharply in 2008 (the Lehman Shock) and in 2020 (the COVID-19 shock).

Fig.~\ref{fig:05_EWS_EDA_monthly_judgment_trend_percentage} is a graph in which the total number of responses in Fig.~\ref{fig:05_EWS_EDA_monthly_judgment_trend} is set to 100\%. The interpretation of this graph is the same as that of Fig.~\ref{fig:05_EWS_EDA_monthly_judgment_trend}.

\subsection{The MLS}\label{subsec:mls}
The MLS, published by the Ministry of Health, Labour and Welfare of Japan, tracks wage trends in Japan, but its publication lags by about two months~\cite{mls}. Given this delay, leveraging the EWS for wage prediction offers practical advantages and allows for both immediate and retrospective analysis, making it possible to track evolving labor market conditions with a higher degree of accuracy. Data are available from January 1990. However, since the EWS data are available only from January 2000 onward as noted in Subsection~\ref{subsec:ews}, only MLS data from January 2000 onward are used in this study.

\begin{figure}[htbp]
	\centering
	
	\begin{minipage}{\linewidth}
		\centering
		\includegraphics[width=\linewidth]{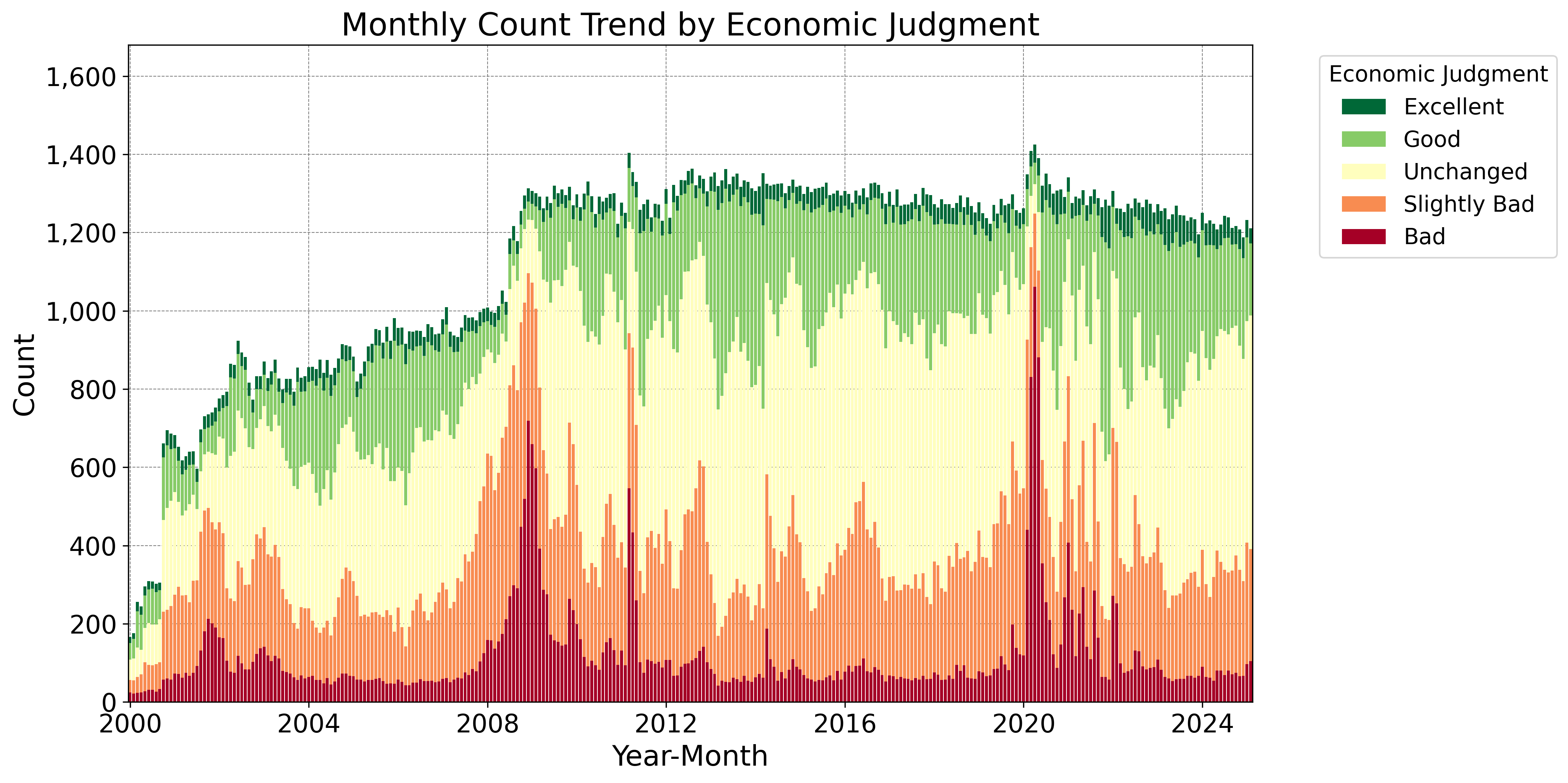}
		\caption{Monthly Count Trend by Economic Judgment}
		\caption*{This figure presents the monthly trend in the number of responses in the EWS survey across Japan from January 2000 to February 2025, along with their breakdown by current economic assessment.}
		\label{fig:05_EWS_EDA_monthly_judgment_trend}
	\end{minipage}
	
	\vspace{1.0em} 
	
	\begin{minipage}{\linewidth}
		\centering
		\includegraphics[width=\linewidth]{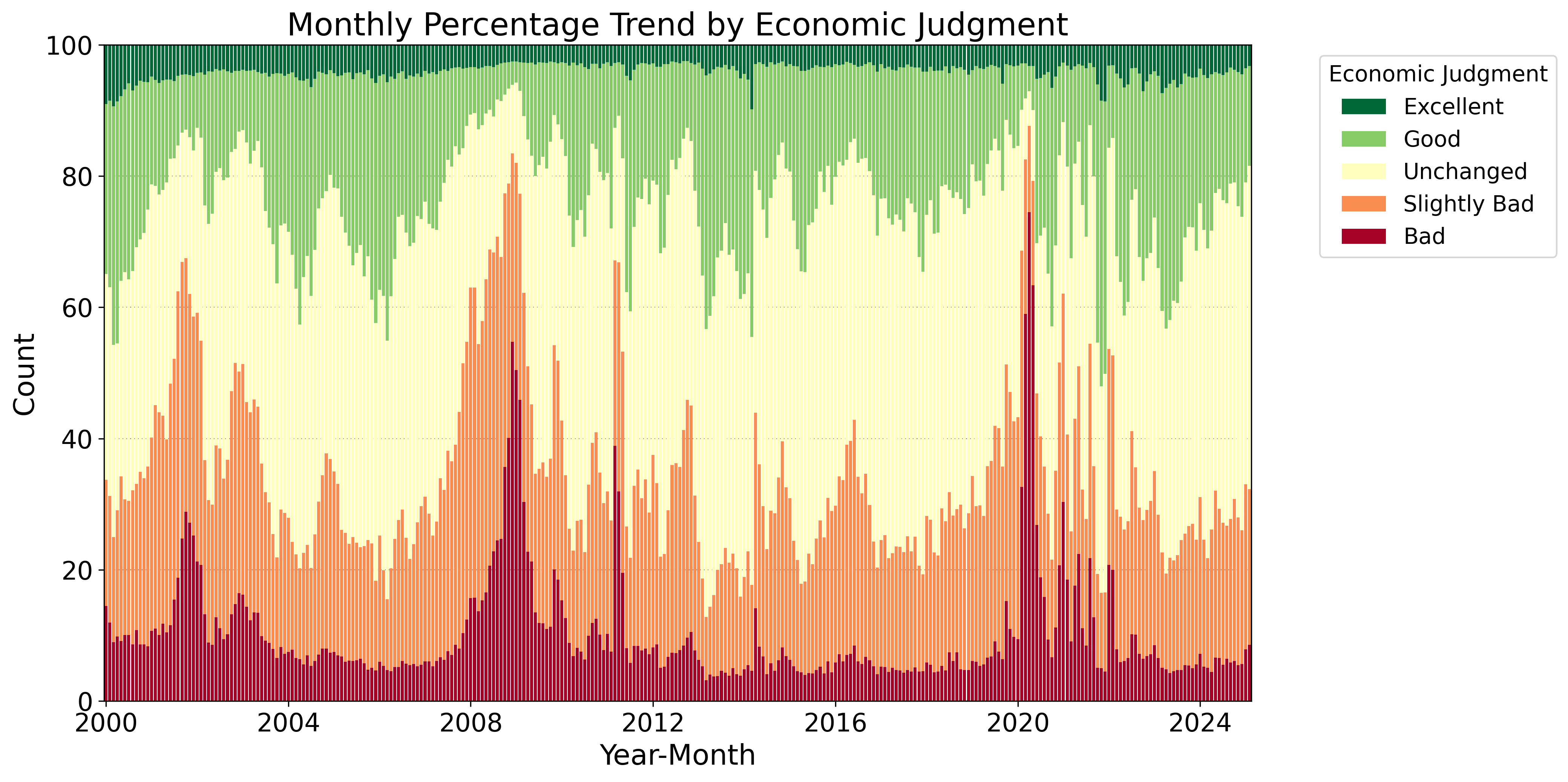}
		\caption{Monthly Percentage Trend by Economic Judgment}
		\caption*{This is a graph in which the total number of responses in Fig.~\ref{fig:05_EWS_EDA_monthly_judgment_trend} is set to 100\%.}
		\label{fig:05_EWS_EDA_monthly_judgment_trend_percentage}
	\end{minipage}
		
	\vspace{1.0em} 	
	
	\begin{minipage}{\linewidth}
		\centering
		\includegraphics[width=\linewidth]{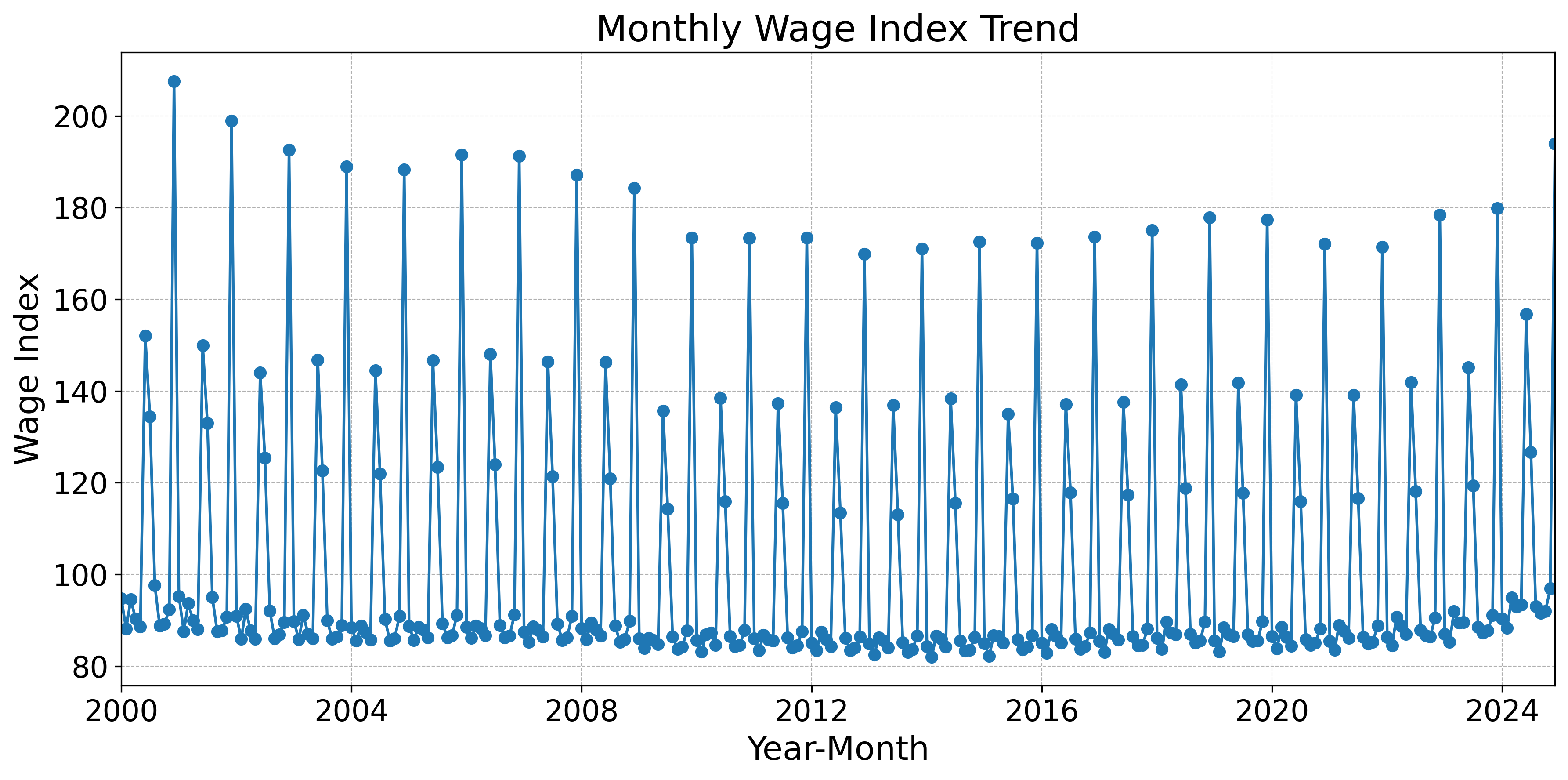}
		\caption{Monthly Wage Index Trend}
		\caption*{This figure presents the monthly series of the nominal wage index from January 2000 to December 2024, derived from the MLS.}
		\label{fig:07_MLS_EDA_yearmonth_trend}
	\end{minipage}
	
	\vspace{1.0em} 
	
	\begin{minipage}{\linewidth}
		\centering
		\includegraphics[width=\linewidth]{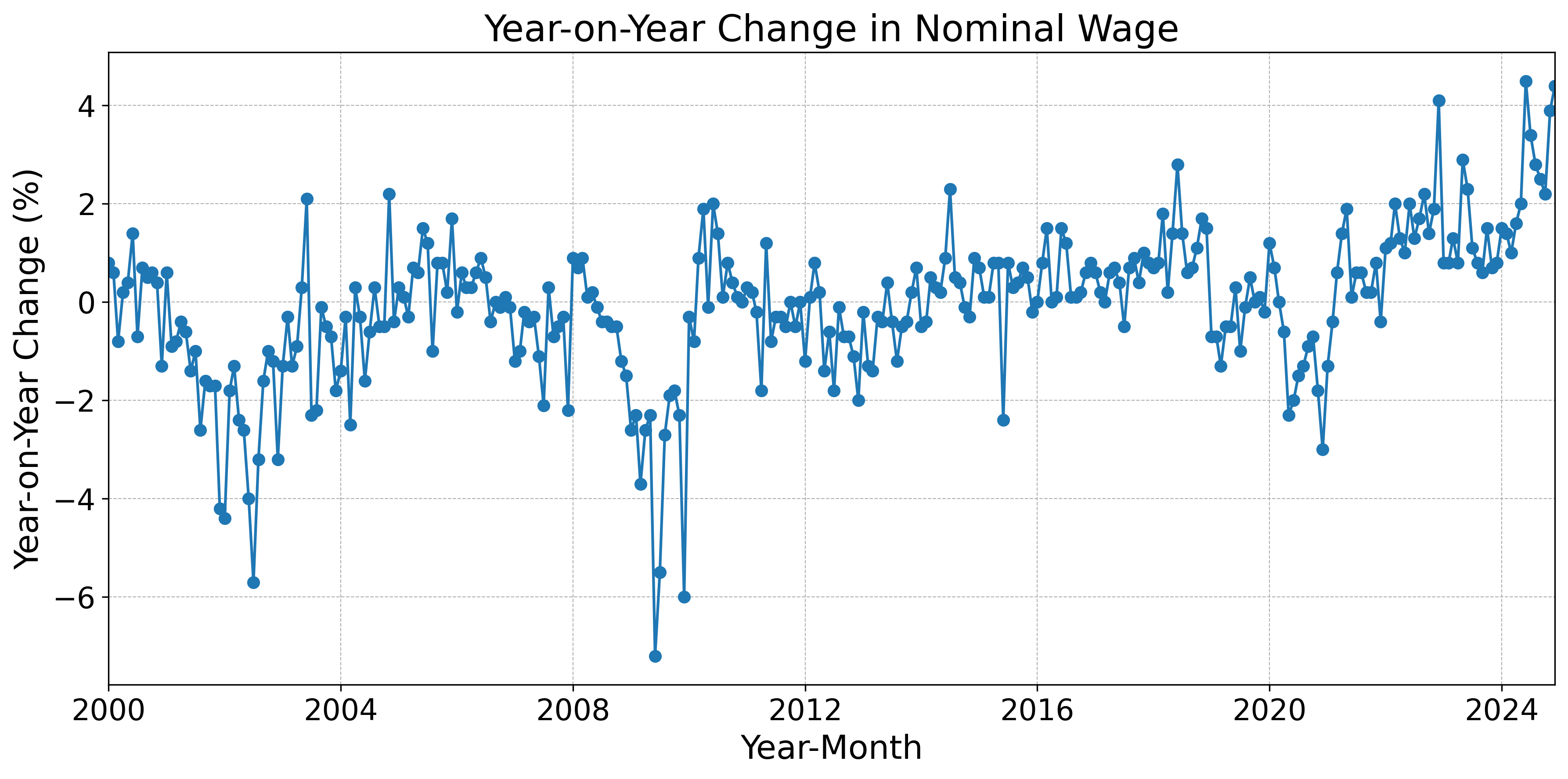}
		\caption{Year-on-Year Change in Wage Index}
		\caption*{This figure presents the monthly series of the year-on-year change in nominal wages from January 2000 to December 2024, derived from the MLS.}
		\label{fig:07_MLS_EDA_yearmonth_yoy_trend}
	\end{minipage}
	
\end{figure}

Fig.~\ref{fig:07_MLS_EDA_yearmonth_trend} shows the monthly series of the nominal wage index from January 2000 to December 2024. The series exhibits a cyclical pattern: wages typically rise in June and July, coinciding with summer bonus payments, and again in December, when winter bonuses are paid.

Fig.~\ref{fig:07_MLS_EDA_yearmonth_yoy_trend} shows the year-on-year change in nominal wages in Japan. The data period is from January 2000 to December 2024. The data show that, after hovering at a low level for several decades, nominal wages have been rising in recent years amid high inflation. Looking back, the negative year-on-year rate of change widened in the early 2000s—when the IT bubble burst and non-regular employment expanded—as well as in 2008 (the Lehman shock) and in 2020 (the COVID-19 shock).

%

\FloatBarrier

\section{Data Architecture}\label{sec:architecture}
\begin{figure*}[htbp]
	\centering
	\includegraphics[width=\textwidth, height=.30625\textheight]{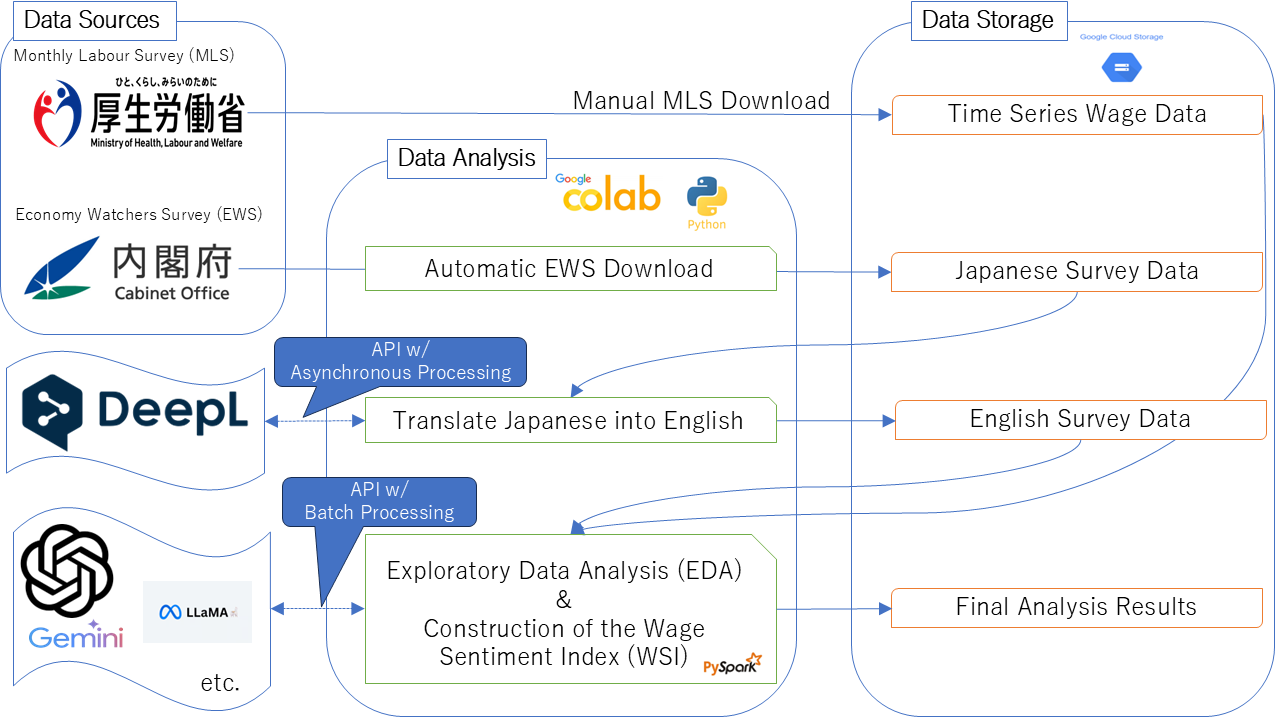}
	\caption{Data Architecture}
	\caption*{This figure illustrates the data architecture constructed in this study. It consists mainly of three components: data sources, data storage, and data analysis. Google Cloud Storage is employed as the data storage. For data analysis, Google Colab is used, and, when necessary, APIs of DeepL and various LLMs are invoked. First, from the Ministry of Health, Labour and Welfare and the Cabinet Office in Japan, the MLS and EWS are obtained, respectively. The former is manually stored in the data storage, while the latter is automatically stored through Colab. The acquired EWS is then translated from Japanese into English. At this stage, the DeepL API is called asynchronously to process the translation efficiently. Next, using the translated EWS together with the MLS, APIs of various LLMs are invoked in batch mode to construct the WSI. Data handling is performed using PySpark. The output results are stored in the data storage. By utilizing this data architecture, scalable and efficient data processing is made possible, with a view to future data expansion.}
	\label{fig:data_architecture}
\end{figure*}

In this study, a scalable data architecture was constructed to enable high-speed processing and accommodate future data expansion. The data architecture is presented in Fig.~\ref{fig:data_architecture}. The data sources consist of MLS and EWS, as described in Section~\ref{sec:data}. MLS has a relatively small data size and is compiled into a single xls file, which is manually uploaded to Google Cloud Storage. In contrast, EWS records the results of current evaluations and future outlook surveys for each month in separate CSV files\footnote{In this study, following the approach of \cite{suzuki2024}, only the analysis results for survey responses related to current evaluations are presented.}. Manually uploading approximately 600 files covering the past 25 years to Google Cloud Storage would be highly labor-intensive. To streamline this process, Python scripts are executed on Google Colab to automate the data upload.

Among the stored MLS and EWS datasets, MLS contains numerical series of past wage indices, whereas EWS consists of Japanese text reporting past economic confidence. To enhance the accuracy of analysis using discriminative pretrained language models and LLMs trained predominantly on English text, the EWS data are translated into English. Given the impracticality of manual translation, Python scripts executed on Google Colab leverage the DeepL API~\cite{deepl_api}, with asynchronous processing implemented to expedite execution. The translated text is stored in Google Cloud Storage.

Subsequently, exploratory data analysis is conducted, and WSIs are constructed using the translated EWS data and the time series of year-on-year changes in nominal wages from MLS. For each model—whether a baseline model, a discriminative pretrained language model, or a LLM—the WSIs are constructed individually, and this procedure is performed for all models, with batch processing employed to minimize API latency, except for the baseline model.

Additionally, managing data frames with PySpark during the analysis process ensures scalability and supports a design capable of accommodating future data growth. The final analysis results are saved to Google Cloud Storage.

In this way, a data architecture capable of high-speed big data processing has been developed and is utilized for analysis.

\section{Results \& Discussions}\label{sec:results}
In this chapter, the time series of the WSIs constructed using each model are plotted in Subsection~\ref{subsec:wts}, and their usefulness as leading indicators of wages is evaluated and compared using the Granger causality test in Subsection~\ref{subsec:gct}. The data period spans from March 2000 to February 2025\footnote{The earliest year and month at which both the EWS and the MLS are available is January 2000. However, in the baseline model, computing the WSI for a given month requires EWS and MLS data from at least two months earlier; consequently, the baseline model can compute the WSI no earlier than March 2000. For consistency, the other models also compute the WSI from March 2000 onward. See Subsubsection~\ref{subsubsec:baseline_model} for details.}.

\subsection{WSI Time Series}\label{subsec:wts}
\begin{figure*}[htbp]
	\centering
	
	\begin{minipage}{0.49\textwidth}
		\centering
		
		\includegraphics[width=\linewidth]{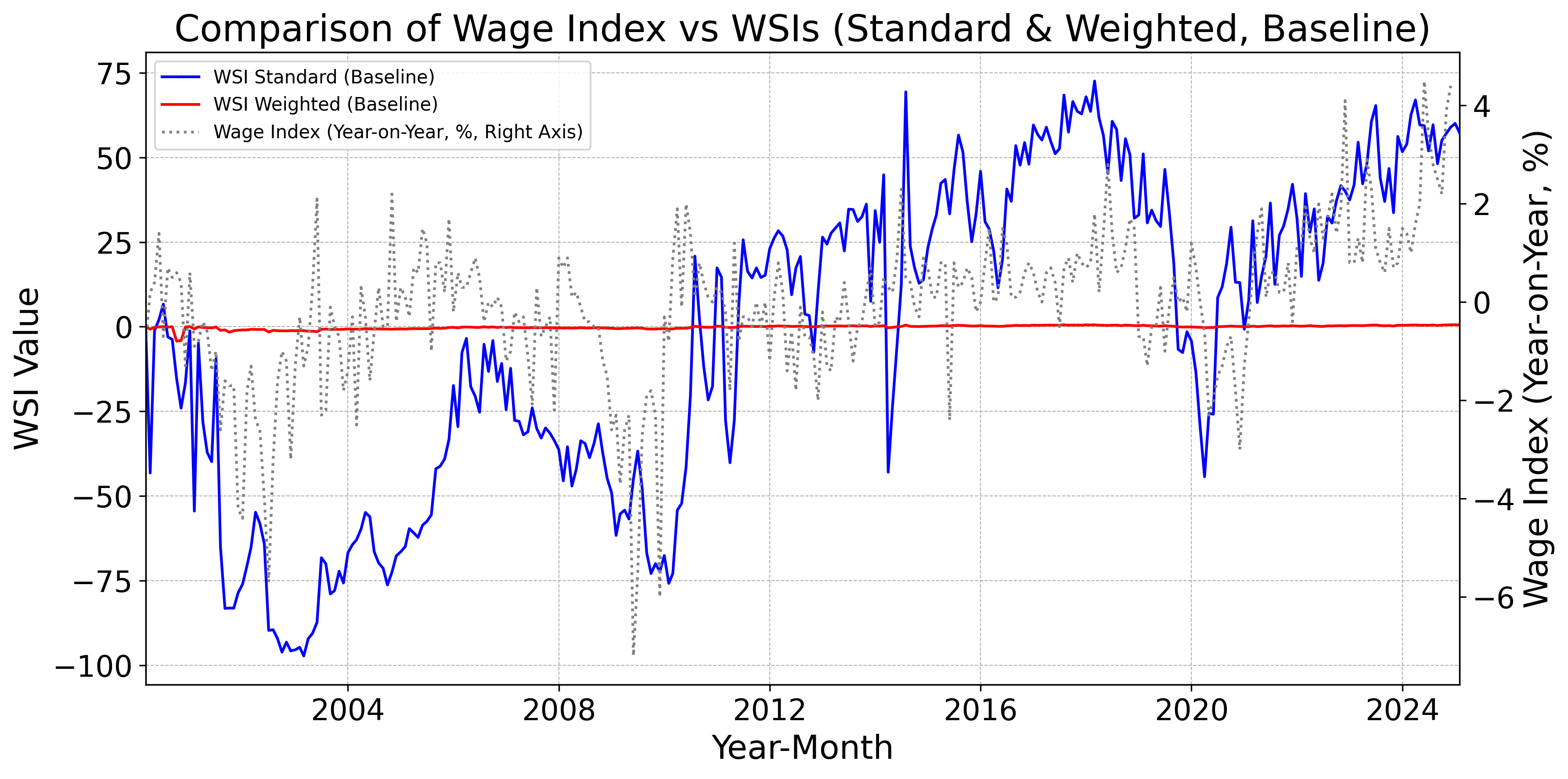}
		\caption{Comparison of Wage Index vs WSIs (Baseline)}
		\caption*{This figure shows the monthly series of the standard and weighted WSIs constructed using the baseline model and the year-on-year change in nominal wages obtained from the MLS. The data period spans from March 2000 to February 2025.}
		\label{fig:08_WSI_Baseline}
		
		\includegraphics[width=\linewidth]{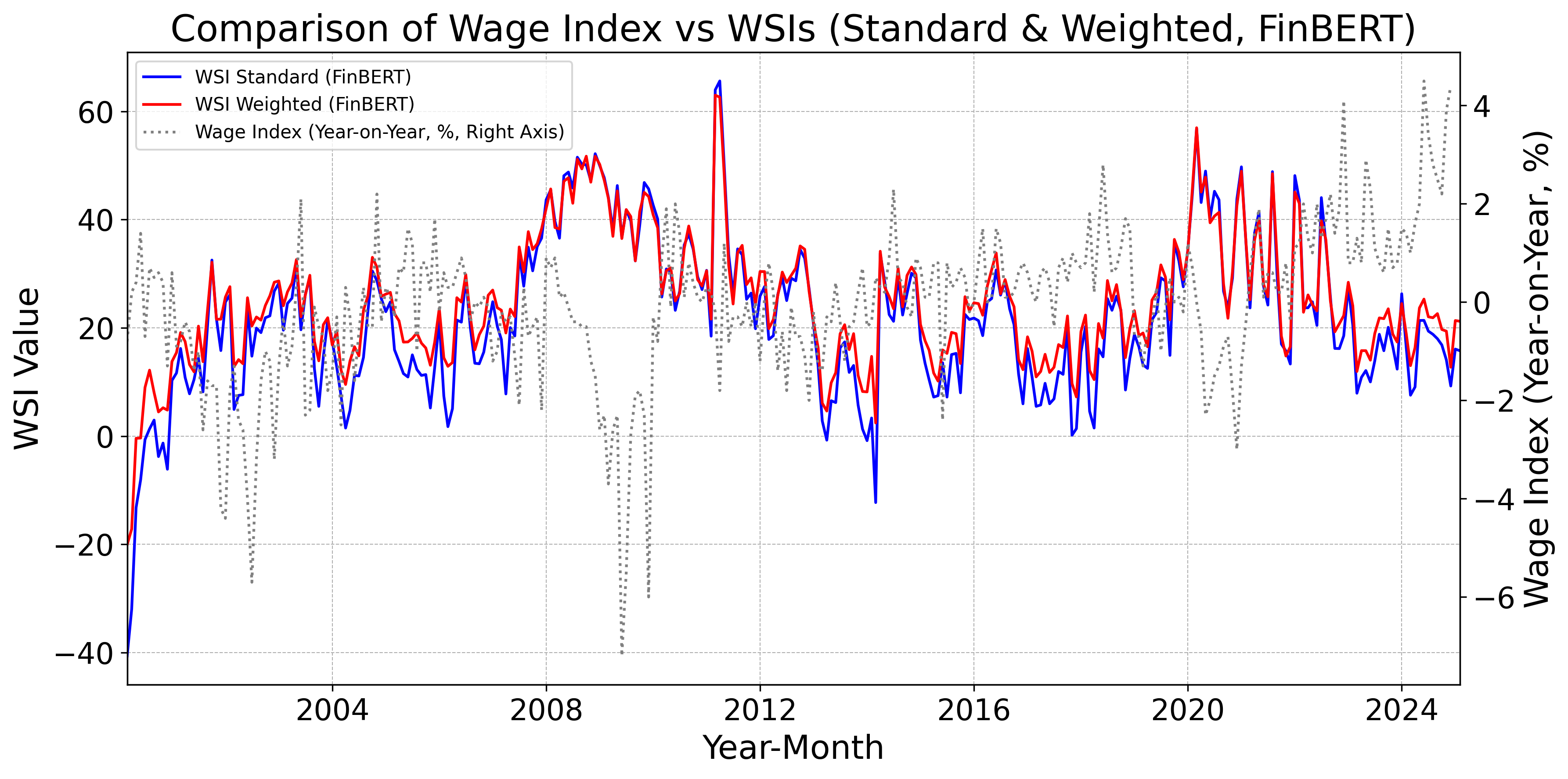}
		\caption{Comparison of Wage Index vs WSIs (FinBERT)}
		\caption*{This figure shows the monthly series of the standard and weighted WSIs constructed using FinBERT and the year-on-year change in nominal wages obtained from the MLS. The data period spans from March 2000 to February 2025.}
		\label{fig:09_WSI_FinBERT}
	\end{minipage}
	\hfill
	\begin{minipage}{0.49\textwidth}
		\centering
		
		\includegraphics[width=\linewidth]{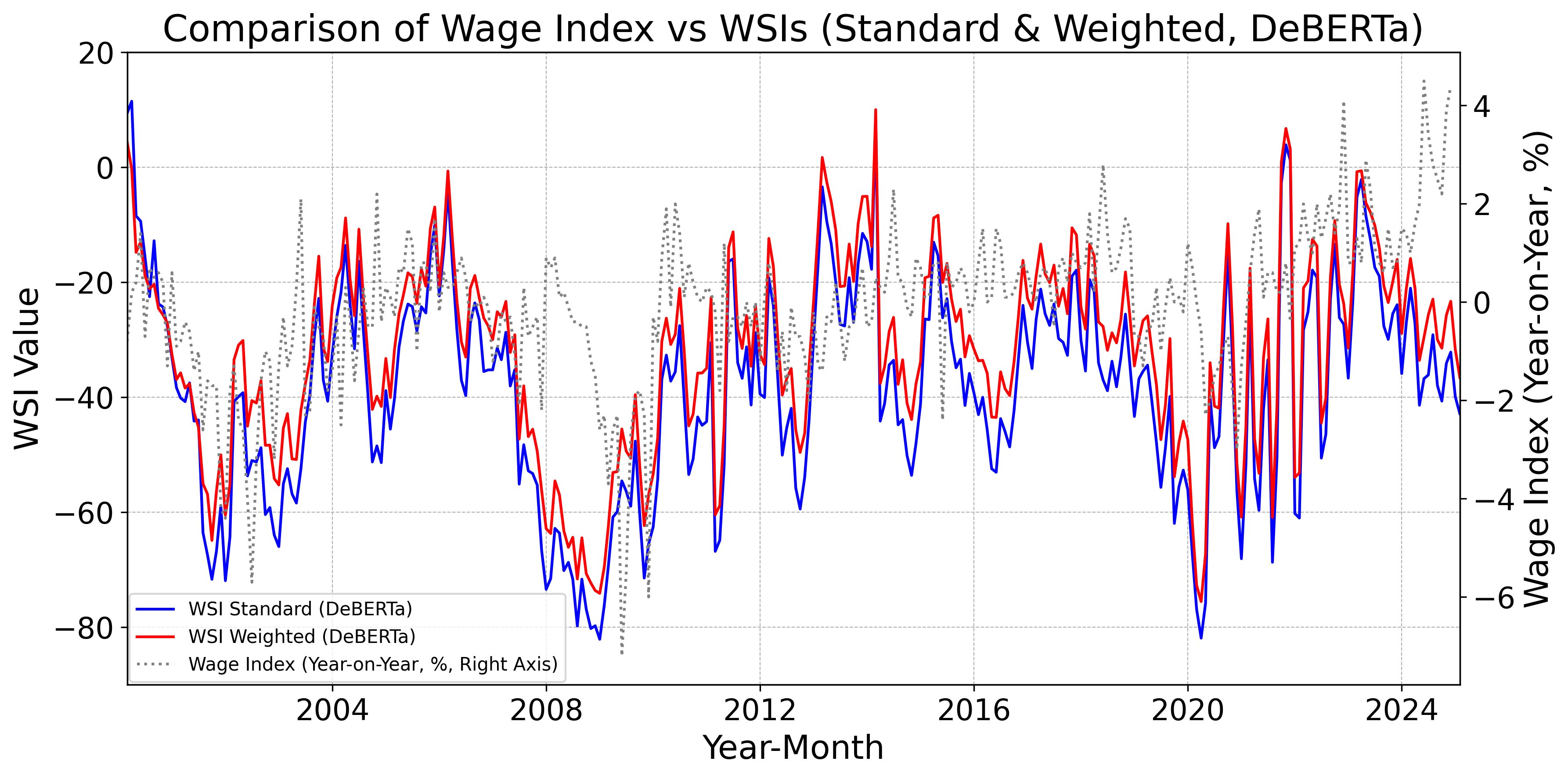}
		\caption{Comparison of Wage Index vs WSIs (DeBERTa)}
		\caption*{This figure shows the monthly series of the standard and weighted WSIs constructed using DeBERTa and the year-on-year change in nominal wages obtained from the MLS. The data period spans from March 2000 to February 2025.}
		\label{fig:10_WSI_DeBERTa}

		\includegraphics[width=\linewidth]{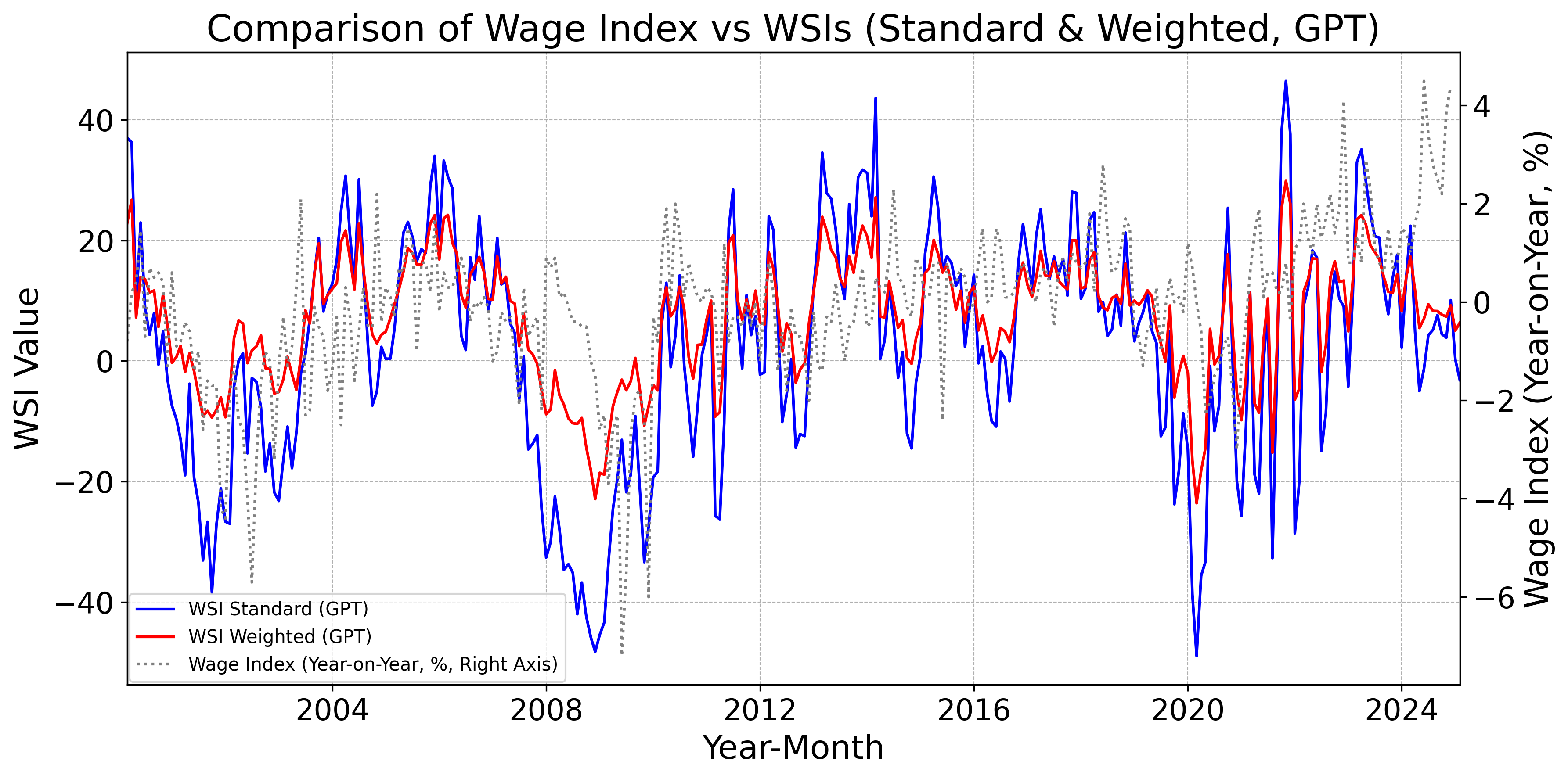}
		\caption{Comparison of Wage Index vs WSIs (GPT)}
		\caption*{This figure shows the monthly series of the standard and weighted WSIs constructed using GPT and the year-on-year change in nominal wages obtained from the MLS. The data period spans from March 2000 to February 2025.\\}
		\label{fig:11_WSI_GPT}

	\end{minipage}
	
\end{figure*}

\begin{figure*}[htbp]
	\centering
	
	\begin{minipage}{0.49\textwidth}
		\centering
		
		\includegraphics[width=\linewidth]{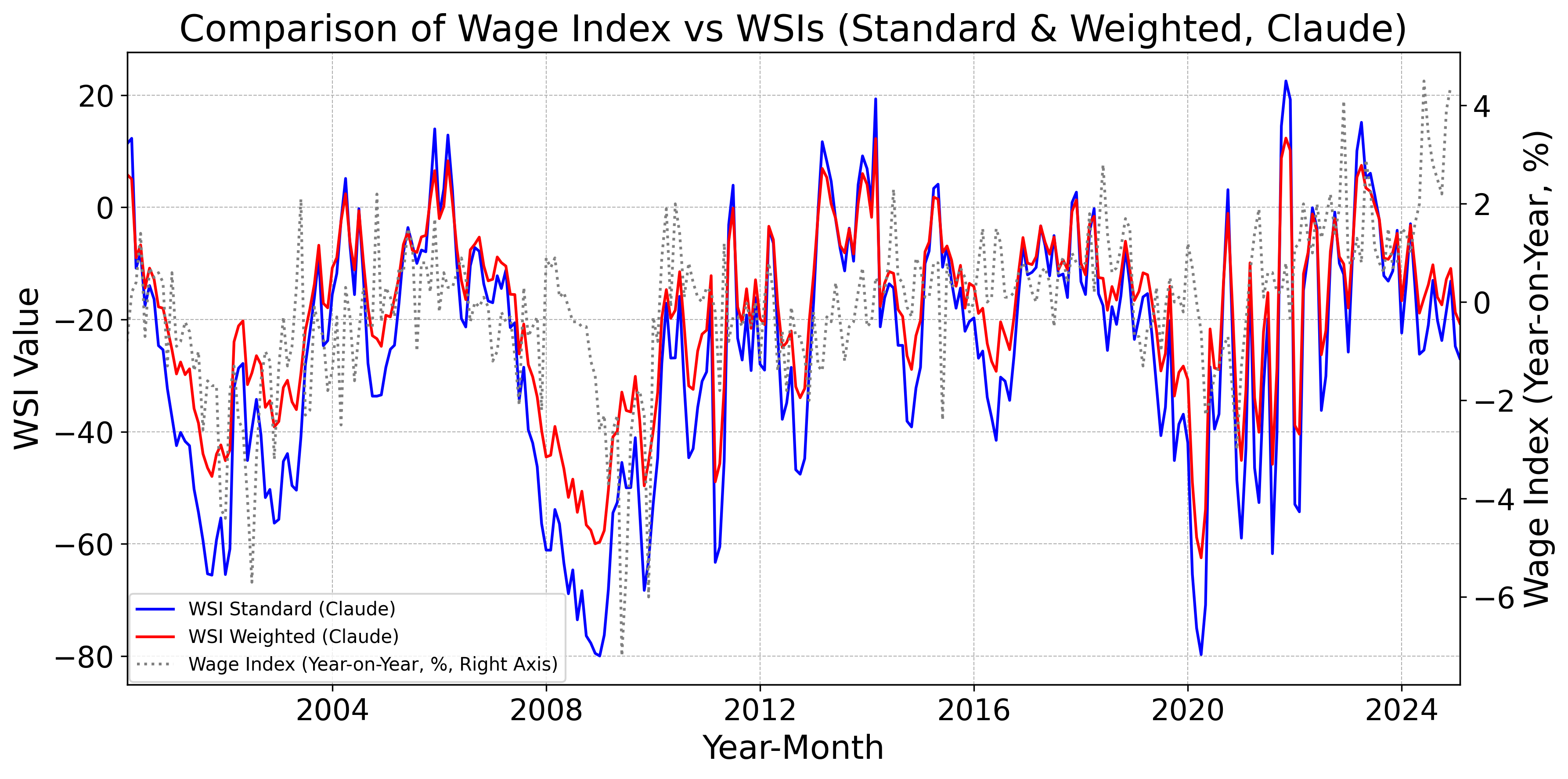}
		\caption{Comparison of Wage Index vs WSIs (Claude)}
		\caption*{This figure shows the monthly series of the standard and weighted WSIs constructed using Claude and the year-on-year change in nominal wages obtained from the MLS. The data period spans from March 2000 to February 2025.}
		\label{fig:12_WSI_Claude}
		
		\includegraphics[width=\linewidth]{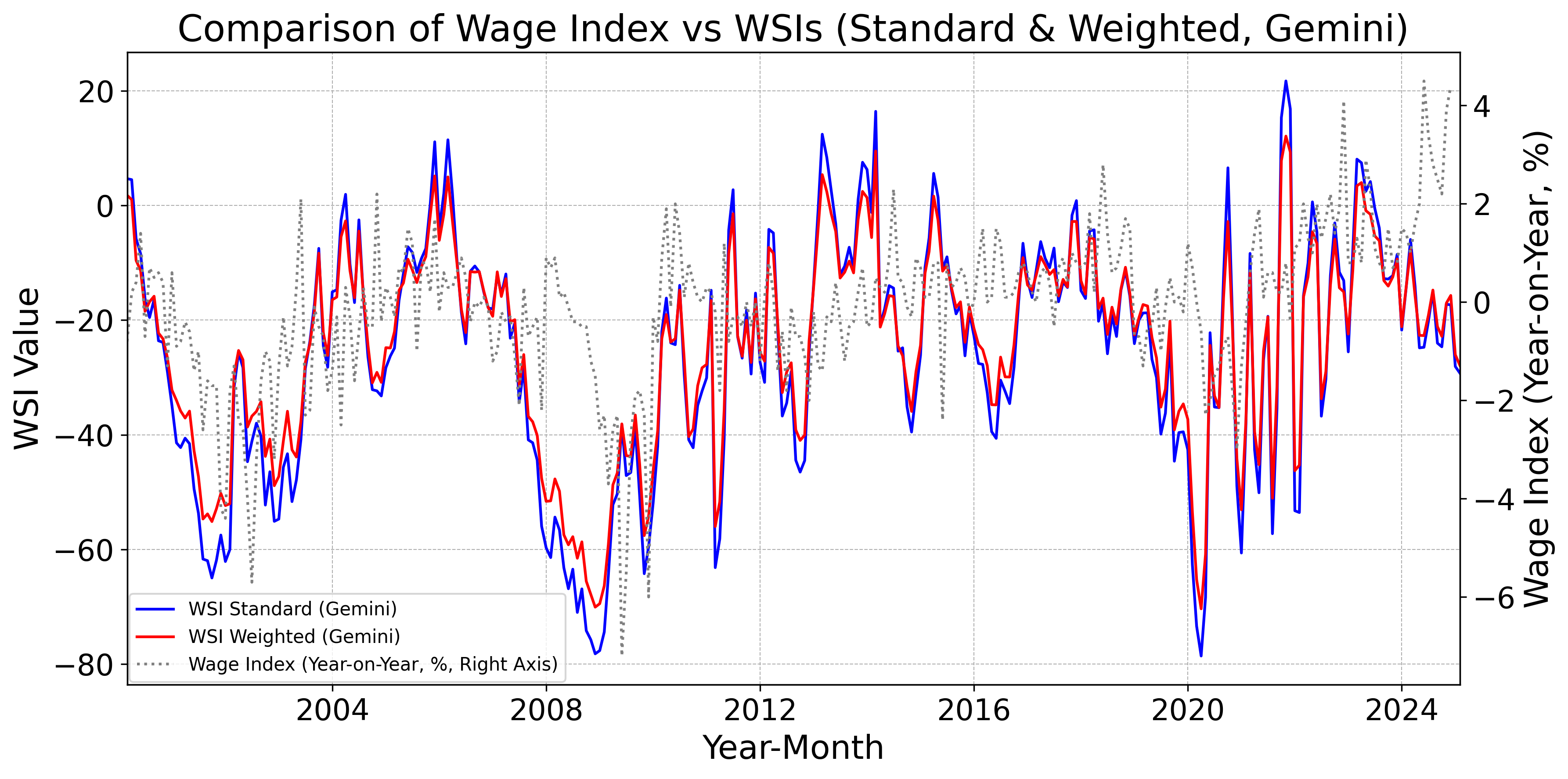}
		\caption{Comparison of Wage Index vs WSIs (Gemini)}
		\caption*{This figure shows the monthly series of the standard and weighted WSIs constructed using Gemini and the year-on-year change in nominal wages obtained from the MLS. The data period spans from March 2000 to February 2025.}
		\label{fig:13_WSI_Gemini}
	\end{minipage}
	\hfill
	\begin{minipage}{0.49\textwidth}
		\centering
		
		\includegraphics[width=\linewidth]{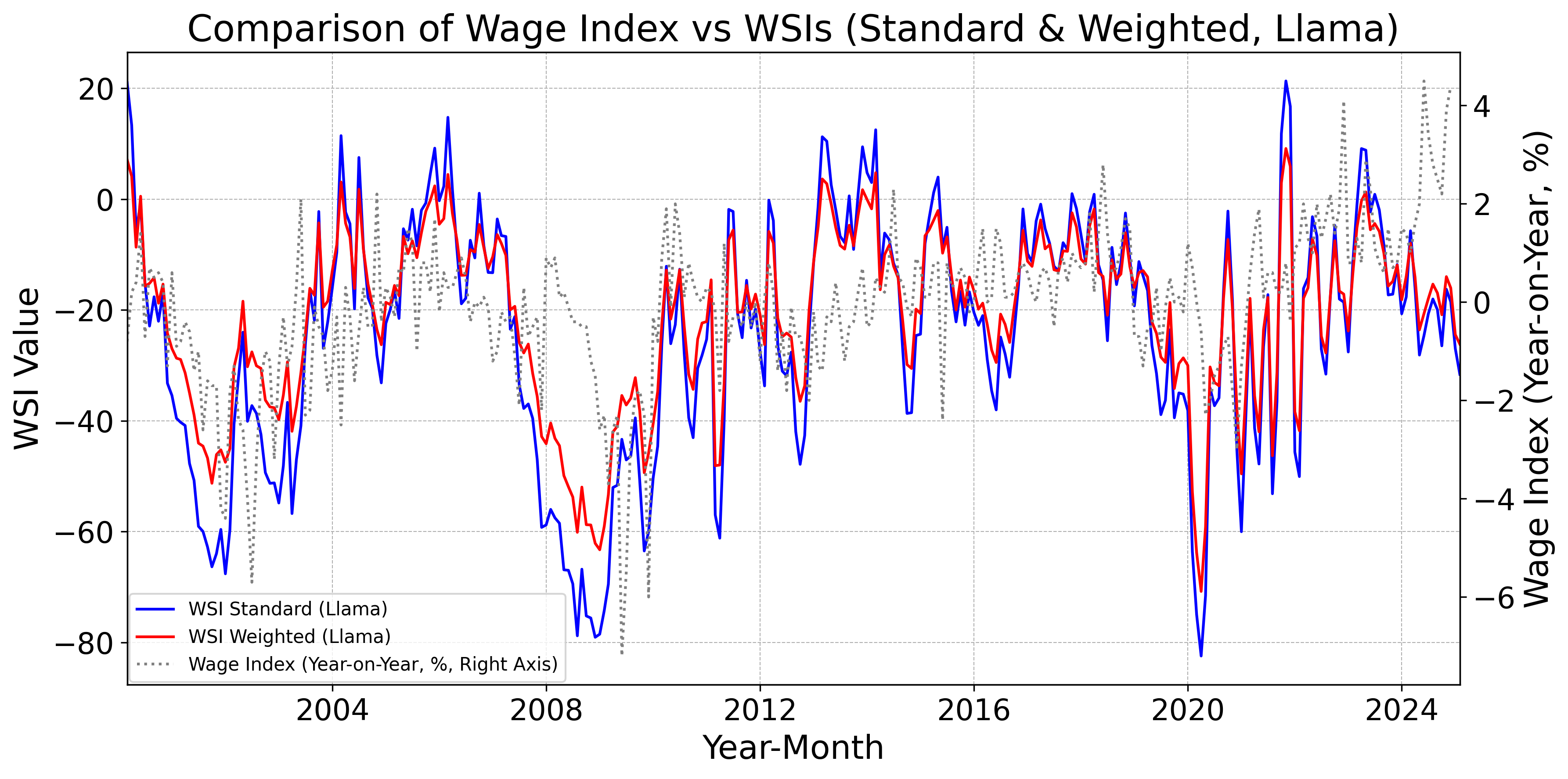}
		\caption{Comparison of Wage Index vs WSIs (Llama)}
		\caption*{This figure shows the monthly series of the standard and weighted WSIs constructed using Llama and the year-on-year change in nominal wages obtained from the MLS. The data period spans from March 2000 to February 2025.}
		\label{fig:14_WSI_Llama}
		
		\includegraphics[width=\linewidth]{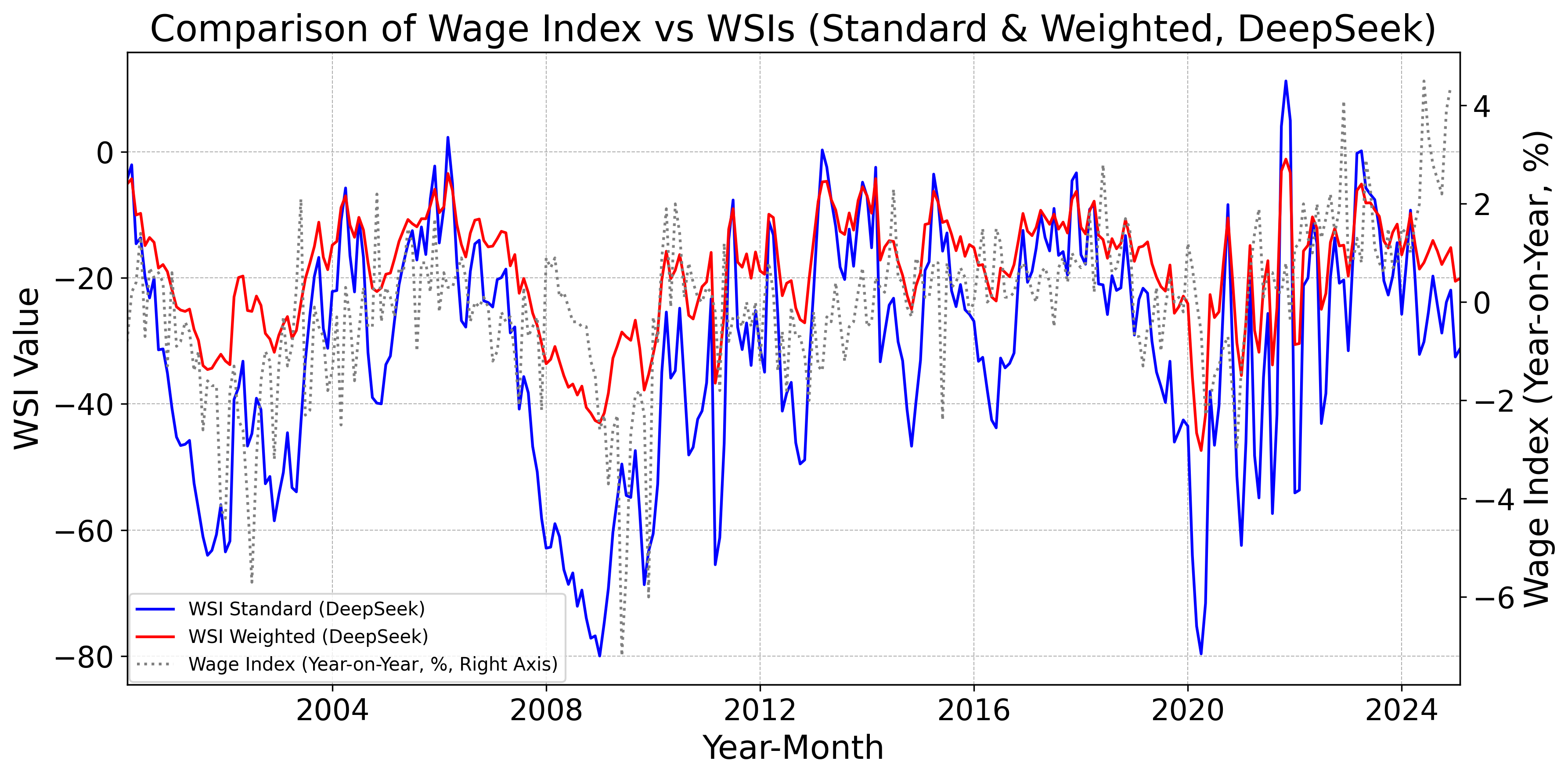}
		\caption{Comparison of Wage Index vs WSIs (DeepSeek)}
		\caption*{This figure shows the monthly series of the standard and weighted WSIs constructed using DeepSeek and the year-on-year change in nominal wages obtained from the MLS. The data period spans from March 2000 to February 2025.}
		\label{fig:15_WSI_DeepSeek}

	\end{minipage}
	
\end{figure*}

Figs.~\ref{fig:08_WSI_Baseline}--\ref{fig:15_WSI_DeepSeek} display, for the baseline model, FinBERT, DeBERTa, GPT, Claude, Gemini, Llama, and DeepSeek, the monthly standard and weighted WSIs alongside the year-on-year change in nominal wages. Fig.~\ref{fig:08_WSI_Baseline} indicates that the baseline WSIs deviate from the year-on-year change in nominal wages. By contrast, Figs.~\ref{fig:09_WSI_FinBERT} and \ref{fig:10_WSI_DeBERTa} show that the DeBERTa-based WSIs track the year-on-year change more closely than the FinBERT-based WSIs. Moreover, Figs.~\ref{fig:11_WSI_GPT}--\ref{fig:15_WSI_DeepSeek} demonstrate that WSIs constructed with LLMs tend to capture the year-on-year change accurately. Taken together, these findings suggest a clear hierarchy: pretrained models tend to outperform the baseline model as leading indicators of wages, and LLM-based WSIs in turn outperform those built with pretrained models.

\subsection{Granger Causality Test}\label{subsec:gct}
Tables~\ref{tab:granger_standard} and~\ref{tab:granger_weighted} show the results of testing whether there is Granger causality between the standard WSI and the weighted WSI constructed using each model and the year-on-year change in nominal wages.

According to Table~\ref{tab:granger_standard}, the standard WSI constructed using LLMs is more useful as a leading indicator of wages than those constructed by the baseline model or by pretrained models. Specifically, when Granger causality tests are applied to the monthly series of the LLM-based standard WSI and the year-on-year change in nominal wages, statistical significance is observed at a greater number of lags. Notably, significance appears at short lags (1–3 months), which are important for policy operations, yielding results consistent with the design objective of compensating for the roughly two-month publication delay of the MLS relative to the EWS.

While Table~\ref{tab:granger_weighted} yields results largely similar to those in Table~\ref{tab:granger_standard}, it differs in that the weighted WSI constructed with DeBERTa exhibits usefulness as a leading indicator of wages comparable to that of the weighted WSIs constructed with LLMs. Specifically, when Granger causality tests are applied to the monthly series of the DeBERTa-based weighted WSI and the year-on-year change in nominal wages, statistical significance is confirmed across a broad range of lags, mirroring the pattern observed when the tests are applied to the LLM-based weighted WSI and nominal wages. This likely reflects the fact that the weighted WSI is more continuous and fine-grained than the standard WSI, aligning well with DeBERTa's capacity to capture subtle contextual nuances. By contrast, FinBERT's class probabilities tend to be overly extreme toward positive or negative outcomes, and, whereas DeBERTa is trained on general-purpose text, FinBERT's training data are skewed toward specialized financial documents; as a result, it is less able to accurately interpret EWS responses from the Japanese general public and, despite also being a pretrained model, fails to match DeBERTa's performance.

In sum, constructing the WSI with LLMs can compensate for the approximately two-month publication lag of the MLS relative to the EWS; used as a leading indicator of wages, it can thereby support the policy operations of the government and the central bank. Moreover, employing a weighted WSI built with DeBERTa in place of LLMs would likely preserve its usefulness as a leading indicator of wages while reducing the budgetary burden by avoiding the API usage costs associated with LLMs.


\begin{table*}[t]
	\centering
	\caption{Granger Causality Test on the Standard WSI}
	\caption*{\justifying\normalfont This table presents the results of the Granger causality tests between the standard WSI constructed by each model and the year-on-year change in nominal wages obtained from the MLS. The null hypothesis states that the standard WSI has no Granger-causal relationship with the year-on-year change in nominal wages. The data period spans from March 2000 to February 2025. The lags range from one month to 24 months. In the table, *, **, and *** denote significance at the 10\%, 5\%, and 1\% levels, respectively.}
	\resizebox{\textwidth}{!}{%
		\begin{tabular}{cccc}
			\textbf{Baseline} &
			\textbf{FinBERT}  &
			\textbf{DeBERTa}  &
			\textbf{GPT} \\
			\begin{tabular}{rrl}
\toprule
Lag & F-stat & p-value \\
\midrule
1 & 18.390 & 0.000*** \\
2 & 5.704 & 0.004*** \\
3 & 2.195 & 0.089* \\
4 & 1.382 & 0.240 \\
5 & 0.870 & 0.502 \\
6 & 1.091 & 0.368 \\
7 & 1.001 & 0.431 \\
8 & 1.212 & 0.292 \\
9 & 1.453 & 0.166 \\
10 & 1.537 & 0.126 \\
11 & 1.529 & 0.121 \\
12 & 1.761 & 0.055* \\
13 & 1.429 & 0.146 \\
14 & 1.394 & 0.156 \\
15 & 1.425 & 0.136 \\
16 & 1.318 & 0.186 \\
17 & 1.257 & 0.222 \\
18 & 1.128 & 0.326 \\
19 & 1.100 & 0.352 \\
20 & 1.016 & 0.444 \\
21 & 0.999 & 0.466 \\
22 & 0.867 & 0.638 \\
23 & 0.905 & 0.592 \\
24 & 0.935 & 0.554 \\
\bottomrule
\end{tabular}
 &
			\begin{tabular}{rrl}
\toprule
Lag & F-stat & p-value \\
\midrule
1 & 1.551 & 0.214 \\
2 & 0.446 & 0.641 \\
3 & 0.827 & 0.480 \\
4 & 0.527 & 0.716 \\
5 & 0.459 & 0.807 \\
6 & 1.116 & 0.353 \\
7 & 1.077 & 0.378 \\
8 & 1.384 & 0.203 \\
9 & 1.137 & 0.337 \\
10 & 1.332 & 0.213 \\
11 & 1.637 & 0.088* \\
12 & 1.403 & 0.164 \\
13 & 1.252 & 0.243 \\
14 & 1.086 & 0.370 \\
15 & 1.671 & 0.057* \\
16 & 1.536 & 0.088* \\
17 & 1.442 & 0.118 \\
18 & 1.416 & 0.124 \\
19 & 1.667 & 0.043** \\
20 & 1.619 & 0.049** \\
21 & 1.542 & 0.065* \\
22 & 1.564 & 0.056* \\
23 & 1.444 & 0.092* \\
24 & 1.551 & 0.054* \\
\bottomrule
\end{tabular}
 &
			\begin{tabular}{rrl}
\toprule
Lag & F-stat & p-value \\
\midrule
1 & 6.332 & 0.012** \\
2 & 2.527 & 0.082* \\
3 & 2.022 & 0.111 \\
4 & 1.287 & 0.275 \\
5 & 1.115 & 0.352 \\
6 & 1.805 & 0.098* \\
7 & 1.635 & 0.125 \\
8 & 1.589 & 0.128 \\
9 & 1.586 & 0.119 \\
10 & 1.649 & 0.093* \\
11 & 1.659 & 0.083* \\
12 & 1.578 & 0.098* \\
13 & 1.378 & 0.170 \\
14 & 1.244 & 0.244 \\
15 & 1.689 & 0.053* \\
16 & 1.625 & 0.063* \\
17 & 1.527 & 0.086* \\
18 & 1.511 & 0.086* \\
19 & 1.468 & 0.098* \\
20 & 1.384 & 0.131 \\
21 & 1.397 & 0.120 \\
22 & 1.245 & 0.212 \\
23 & 1.241 & 0.212 \\
24 & 1.431 & 0.094* \\
\bottomrule
\end{tabular}
 &
			\begin{tabular}{rrl}
\toprule
Lag & F-stat & p-value \\
\midrule
1 & 6.342 & 0.012** \\
2 & 2.719 & 0.068* \\
3 & 2.288 & 0.079* \\
4 & 1.727 & 0.144 \\
5 & 1.521 & 0.183 \\
6 & 2.186 & 0.044** \\
7 & 1.890 & 0.071* \\
8 & 1.778 & 0.081* \\
9 & 1.850 & 0.060* \\
10 & 1.803 & 0.060* \\
11 & 1.963 & 0.032** \\
12 & 1.824 & 0.045** \\
13 & 1.641 & 0.074* \\
14 & 1.469 & 0.123 \\
15 & 1.781 & 0.038** \\
16 & 1.643 & 0.059* \\
17 & 1.594 & 0.066* \\
18 & 1.610 & 0.058* \\
19 & 1.549 & 0.071* \\
20 & 1.459 & 0.097* \\
21 & 1.354 & 0.143 \\
22 & 1.399 & 0.116 \\
23 & 1.348 & 0.138 \\
24 & 1.649 & 0.033** \\
\bottomrule
\end{tabular}
 \\\\
			\textbf{Claude} &
			\textbf{Gemini} &
			\textbf{Llama} &
			\textbf{DeepSeek} \\
			\begin{tabular}{rrl}
\toprule
Lag & F-stat & p-value \\
\midrule
1 & 9.683 & 0.002*** \\
2 & 3.672 & 0.027** \\
3 & 3.278 & 0.021** \\
4 & 2.259 & 0.063* \\
5 & 1.830 & 0.107 \\
6 & 2.411 & 0.027** \\
7 & 2.136 & 0.040** \\
8 & 2.057 & 0.040** \\
9 & 2.053 & 0.034** \\
10 & 2.001 & 0.034** \\
11 & 2.089 & 0.021** \\
12 & 2.025 & 0.022** \\
13 & 1.711 & 0.059* \\
14 & 1.548 & 0.094* \\
15 & 1.922 & 0.022** \\
16 & 1.825 & 0.028** \\
17 & 1.737 & 0.037** \\
18 & 1.641 & 0.051* \\
19 & 1.612 & 0.054* \\
20 & 1.524 & 0.074* \\
21 & 1.497 & 0.079* \\
22 & 1.410 & 0.111 \\
23 & 1.312 & 0.161 \\
24 & 1.432 & 0.094* \\
\bottomrule
\end{tabular}
 &
			\begin{tabular}{rrl}
\toprule
Lag & F-stat & p-value \\
\midrule
1 & 9.255 & 0.003*** \\
2 & 3.370 & 0.036** \\
3 & 3.738 & 0.012** \\
4 & 2.576 & 0.038** \\
5 & 2.056 & 0.071* \\
6 & 2.592 & 0.018** \\
7 & 2.323 & 0.026** \\
8 & 2.232 & 0.025** \\
9 & 2.190 & 0.023** \\
10 & 2.122 & 0.023** \\
11 & 2.196 & 0.015** \\
12 & 2.139 & 0.015** \\
13 & 1.823 & 0.040** \\
14 & 1.633 & 0.071* \\
15 & 1.967 & 0.018** \\
16 & 1.851 & 0.026** \\
17 & 1.751 & 0.035** \\
18 & 1.656 & 0.048** \\
19 & 1.625 & 0.051* \\
20 & 1.569 & 0.061* \\
21 & 1.543 & 0.065* \\
22 & 1.462 & 0.088* \\
23 & 1.404 & 0.110 \\
24 & 1.590 & 0.044** \\
\bottomrule
\end{tabular}
 &
			\begin{tabular}{rrl}
\toprule
Lag & F-stat & p-value \\
\midrule
1 & 10.145 & 0.002*** \\
2 & 3.724 & 0.025** \\
3 & 3.258 & 0.022** \\
4 & 2.209 & 0.068* \\
5 & 1.784 & 0.116 \\
6 & 2.236 & 0.040** \\
7 & 1.965 & 0.060* \\
8 & 1.902 & 0.060* \\
9 & 1.901 & 0.052* \\
10 & 1.886 & 0.047** \\
11 & 1.869 & 0.044** \\
12 & 1.739 & 0.059* \\
13 & 1.536 & 0.105 \\
14 & 1.408 & 0.149 \\
15 & 1.555 & 0.087* \\
16 & 1.418 & 0.133 \\
17 & 1.338 & 0.169 \\
18 & 1.274 & 0.206 \\
19 & 1.215 & 0.245 \\
20 & 1.167 & 0.284 \\
21 & 1.210 & 0.244 \\
22 & 1.053 & 0.401 \\
23 & 0.992 & 0.477 \\
24 & 1.166 & 0.275 \\
\bottomrule
\end{tabular}
 &
			\begin{tabular}{rrl}
\toprule
Lag & F-stat & p-value \\
\midrule
1 & 10.011 & 0.002*** \\
2 & 3.679 & 0.026** \\
3 & 3.901 & 0.009*** \\
4 & 2.637 & 0.034** \\
5 & 2.341 & 0.042** \\
6 & 2.520 & 0.022** \\
7 & 2.258 & 0.030** \\
8 & 2.200 & 0.028** \\
9 & 2.138 & 0.027** \\
10 & 2.145 & 0.022** \\
11 & 2.184 & 0.016** \\
12 & 2.088 & 0.018** \\
13 & 1.721 & 0.057* \\
14 & 1.621 & 0.074* \\
15 & 1.749 & 0.042** \\
16 & 1.620 & 0.064* \\
17 & 1.535 & 0.083* \\
18 & 1.482 & 0.097* \\
19 & 1.424 & 0.116 \\
20 & 1.335 & 0.158 \\
21 & 1.395 & 0.121 \\
22 & 1.264 & 0.198 \\
23 & 1.188 & 0.257 \\
24 & 1.338 & 0.141 \\
\bottomrule
\end{tabular}
 \\
		\end{tabular}
	}
	\label{tab:granger_standard}
\end{table*}


\begin{table*}[t]
	\centering
	\caption{Granger Causality Test on the Weighted WSI}
	\caption*{\justifying\normalfont This table presents the results of the Granger causality tests between the weighted WSI constructed by each model and the year-on-year change in nominal wages obtained from the MLS. The null hypothesis states that the weighted WSI has no Granger-causal relationship with the year-on-year change in nominal wages. The data period spans from March 2000 to February 2025. The lags range from one month to 24 months. In the table, *, **, and *** denote significance at the 10\%, 5\%, and 1\% levels, respectively.}
	\resizebox{\textwidth}{!}{%
		\begin{tabular}{cccc}
			\textbf{Baseline} &
			\textbf{FinBERT}  &
			\textbf{DeBERTa}  &
			\textbf{GPT} \\
			\begin{tabular}{rrl}
\toprule
Lag & F-stat & p-value \\
\midrule
1 & 14.713 & 0.000*** \\
2 & 5.077 & 0.007*** \\
3 & 2.342 & 0.073* \\
4 & 1.626 & 0.168 \\
5 & 0.982 & 0.429 \\
6 & 0.994 & 0.430 \\
7 & 1.005 & 0.428 \\
8 & 1.265 & 0.262 \\
9 & 1.409 & 0.184 \\
10 & 1.152 & 0.324 \\
11 & 1.185 & 0.297 \\
12 & 1.188 & 0.292 \\
13 & 1.084 & 0.373 \\
14 & 1.117 & 0.343 \\
15 & 1.538 & 0.092* \\
16 & 1.538 & 0.087* \\
17 & 1.602 & 0.064* \\
18 & 1.529 & 0.081* \\
19 & 1.590 & 0.059* \\
20 & 1.965 & 0.010*** \\
21 & 1.885 & 0.013** \\
22 & 1.252 & 0.206 \\
23 & 1.291 & 0.175 \\
24 & 1.353 & 0.133 \\
\bottomrule
\end{tabular}
 &
			\begin{tabular}{rrl}
\toprule
Lag & F-stat & p-value \\
\midrule
1 & 1.610 & 0.205 \\
2 & 0.453 & 0.636 \\
3 & 0.831 & 0.478 \\
4 & 0.514 & 0.725 \\
5 & 0.459 & 0.807 \\
6 & 0.811 & 0.562 \\
7 & 0.747 & 0.632 \\
8 & 1.059 & 0.392 \\
9 & 0.880 & 0.543 \\
10 & 1.090 & 0.370 \\
11 & 1.580 & 0.104 \\
12 & 1.342 & 0.195 \\
13 & 1.223 & 0.263 \\
14 & 1.048 & 0.406 \\
15 & 1.635 & 0.065* \\
16 & 1.543 & 0.085* \\
17 & 1.452 & 0.114 \\
18 & 1.433 & 0.116 \\
19 & 1.643 & 0.047** \\
20 & 1.562 & 0.063* \\
21 & 1.514 & 0.074* \\
22 & 1.607 & 0.046** \\
23 & 1.482 & 0.078* \\
24 & 1.634 & 0.036** \\
\bottomrule
\end{tabular}
 &
			\begin{tabular}{rrl}
\toprule
Lag & F-stat & p-value \\
\midrule
1 & 6.794 & 0.010*** \\
2 & 2.706 & 0.068* \\
3 & 2.675 & 0.047** \\
4 & 1.772 & 0.134 \\
5 & 1.494 & 0.192 \\
6 & 2.310 & 0.034** \\
7 & 2.063 & 0.048** \\
8 & 1.974 & 0.050** \\
9 & 1.932 & 0.048** \\
10 & 1.950 & 0.039** \\
11 & 1.905 & 0.039** \\
12 & 1.792 & 0.050** \\
13 & 1.535 & 0.105 \\
14 & 1.353 & 0.177 \\
15 & 1.910 & 0.023** \\
16 & 1.831 & 0.028** \\
17 & 1.721 & 0.040** \\
18 & 1.681 & 0.043** \\
19 & 1.616 & 0.053* \\
20 & 1.529 & 0.073* \\
21 & 1.531 & 0.068* \\
22 & 1.470 & 0.085* \\
23 & 1.425 & 0.100 \\
24 & 1.606 & 0.041** \\
\bottomrule
\end{tabular}
 &
			\begin{tabular}{rrl}
\toprule
Lag & F-stat & p-value \\
\midrule
1 & 8.699 & 0.003*** \\
2 & 3.467 & 0.032** \\
3 & 2.737 & 0.044** \\
4 & 2.046 & 0.088* \\
5 & 1.669 & 0.142 \\
6 & 2.594 & 0.018** \\
7 & 2.221 & 0.033** \\
8 & 2.139 & 0.033** \\
9 & 2.055 & 0.034** \\
10 & 1.987 & 0.035** \\
11 & 1.957 & 0.033** \\
12 & 1.784 & 0.051* \\
13 & 1.537 & 0.104 \\
14 & 1.403 & 0.152 \\
15 & 1.772 & 0.039** \\
16 & 1.658 & 0.055* \\
17 & 1.603 & 0.064* \\
18 & 1.580 & 0.066* \\
19 & 1.482 & 0.092* \\
20 & 1.402 & 0.122 \\
21 & 1.416 & 0.111 \\
22 & 1.372 & 0.129 \\
23 & 1.307 & 0.164 \\
24 & 1.494 & 0.070* \\
\bottomrule
\end{tabular}
 \\\\
			\textbf{Claude} &
			\textbf{Gemini} &
			\textbf{Llama} &
			\textbf{DeepSeek} \\
			\begin{tabular}{rrl}
\toprule
Lag & F-stat & p-value \\
\midrule
1 & 10.548 & 0.001*** \\
2 & 3.908 & 0.021** \\
3 & 3.705 & 0.012** \\
4 & 2.584 & 0.037** \\
5 & 2.042 & 0.073* \\
6 & 2.427 & 0.026** \\
7 & 2.168 & 0.037** \\
8 & 2.086 & 0.037** \\
9 & 2.041 & 0.035** \\
10 & 1.984 & 0.035** \\
11 & 2.066 & 0.023** \\
12 & 1.956 & 0.029** \\
13 & 1.663 & 0.069* \\
14 & 1.496 & 0.112 \\
15 & 1.876 & 0.026** \\
16 & 1.763 & 0.037** \\
17 & 1.687 & 0.046** \\
18 & 1.629 & 0.054* \\
19 & 1.567 & 0.065* \\
20 & 1.492 & 0.085* \\
21 & 1.519 & 0.072* \\
22 & 1.420 & 0.106 \\
23 & 1.346 & 0.140 \\
24 & 1.462 & 0.082* \\
\bottomrule
\end{tabular}
 &
			\begin{tabular}{rrl}
\toprule
Lag & F-stat & p-value \\
\midrule
1 & 10.554 & 0.001*** \\
2 & 3.838 & 0.023** \\
3 & 3.851 & 0.010*** \\
4 & 2.650 & 0.034** \\
5 & 2.056 & 0.071* \\
6 & 2.518 & 0.022** \\
7 & 2.249 & 0.031** \\
8 & 2.184 & 0.029** \\
9 & 2.105 & 0.029** \\
10 & 2.060 & 0.028** \\
11 & 2.097 & 0.021** \\
12 & 1.992 & 0.025** \\
13 & 1.700 & 0.061* \\
14 & 1.542 & 0.097* \\
15 & 1.832 & 0.031** \\
16 & 1.707 & 0.046** \\
17 & 1.624 & 0.059* \\
18 & 1.556 & 0.072* \\
19 & 1.493 & 0.088* \\
20 & 1.430 & 0.109 \\
21 & 1.455 & 0.095* \\
22 & 1.398 & 0.116 \\
23 & 1.364 & 0.130 \\
24 & 1.537 & 0.058* \\
\bottomrule
\end{tabular}
 &
			\begin{tabular}{rrl}
\toprule
Lag & F-stat & p-value \\
\midrule
1 & 10.286 & 0.001*** \\
2 & 3.865 & 0.022** \\
3 & 3.160 & 0.025** \\
4 & 2.169 & 0.073* \\
5 & 1.830 & 0.107 \\
6 & 1.975 & 0.069* \\
7 & 1.767 & 0.094* \\
8 & 1.695 & 0.099* \\
9 & 1.655 & 0.100* \\
10 & 1.706 & 0.079* \\
11 & 1.684 & 0.077* \\
12 & 1.572 & 0.100* \\
13 & 1.388 & 0.165 \\
14 & 1.290 & 0.213 \\
15 & 1.526 & 0.096* \\
16 & 1.394 & 0.145 \\
17 & 1.338 & 0.170 \\
18 & 1.312 & 0.181 \\
19 & 1.249 & 0.219 \\
20 & 1.187 & 0.266 \\
21 & 1.215 & 0.239 \\
22 & 1.110 & 0.337 \\
23 & 1.050 & 0.404 \\
24 & 1.189 & 0.254 \\
\bottomrule
\end{tabular}
 & 
			\begin{tabular}{rrl}
\toprule
Lag & F-stat & p-value \\
\midrule
1 & 11.196 & 0.001*** \\
2 & 4.026 & 0.019** \\
3 & 3.955 & 0.009*** \\
4 & 2.723 & 0.030** \\
5 & 2.265 & 0.048** \\
6 & 2.229 & 0.041** \\
7 & 2.016 & 0.053* \\
8 & 1.979 & 0.049** \\
9 & 1.884 & 0.054* \\
10 & 1.906 & 0.044** \\
11 & 1.995 & 0.029** \\
12 & 1.855 & 0.040** \\
13 & 1.569 & 0.094* \\
14 & 1.463 & 0.125 \\
15 & 1.748 & 0.043** \\
16 & 1.632 & 0.061* \\
17 & 1.568 & 0.073* \\
18 & 1.563 & 0.070* \\
19 & 1.500 & 0.086* \\
20 & 1.432 & 0.108 \\
21 & 1.570 & 0.057* \\
22 & 1.419 & 0.106 \\
23 & 1.355 & 0.135 \\
24 & 1.457 & 0.083* \\
\bottomrule
\end{tabular}
 \\
		\end{tabular}
	}
	\label{tab:granger_weighted}
\end{table*}

\section{Conclusion}\label{sec:conclusion}
In this study, WSIs were constructed from text data using ML models to enable rapid assessment of wage trends. Using LLMs, wage dynamics were captured accurately and promptly. The findings further suggest that the use of the particular pretrained model, in place of LLMs, may offer a viable means of reducing the cost associated with constructing WSIs. To accommodate data sources beyond the EWS employed here (e.g., social media and newspapers) and to enable high-speed, large-scale processing, a scalable data architecture was also developed. These results can help governments and central banks analyze large volumes of data, promptly track wage trends, and design systems and policies.

Future work consists of both short- and longer-term objectives. In the short term, the focus lies on fine-tuning LLMs to create hawkish and dovish variants, as prior research~\cite{duan2025} has demonstrated that model characteristics affect performance. Additional tasks include evaluating changes in the accuracy of price and wage forecasts, as well as incorporating other text sources such as newspapers and social media, combined with ensemble modeling, to improve robustness. In the longer term, the aim is to construct a more scalable and high-speed data pipeline, embedding AI agents capable of managing increasingly diverse and voluminous data, thereby providing strong support for the timely and appropriate policy formulation of governments and central banks.

\section*{Notes}\label{sec:notes}
The opinions expressed in this article belong to the author alone and do not represent the official views of his affiliated institution. Additionally, all possible errors are solely the author's own.

\section*{Acknowledgment}\label{sec:acknowledgments}
This research was conducted at the University of Colorado Boulder. I am grateful to Professor Qin ``Christine'' Lv, Mr. Rahul Aedula, and the fellow students for their insightful feedback and support.

\FloatBarrier

\bibliographystyle{IEEEtran}
\bibliography{mybib_IEEEfull}



\end{document}